\documentclass[a4paper,fleqn,review]{cas-dc}
\usepackage{lscape}
\usepackage{pdflscape}
\usepackage{graphicx}
\usepackage{caption}
\usepackage{subcaption}
\usepackage{amsmath}
\usepackage{array} 

\usepackage[authoryear]{natbib}
\def\tsc#1{\csdef{#1}{\textsc{\lowercase{#1}}\xspace}}
\tsc{WGM}
\tsc{QE}

\begin{document}
\let\WriteBookmarks\relax
\def\floatpagepagefraction{1}
\def\textpagefraction{.001}
\makeatletter\def\Hy@Warning#1{}\makeatother
\shorttitle{Advances in Medical Image Segmentation: A Comprehensive Survey}    

\shortauthors{A. Kabil et. al}  

\title [mode = title]{Advances in Medical Image Segmentation: A Comprehensive Survey with a Focus on Lumbar Spine Applications}  



\author[1]{Ahmed Kabil}



\ead{a.mohamad2116@nu.edu.eg}



\affiliation[1]{organization={Center for Informatics Science, School of Information Technology and Computer Science, Nile University},
            postcode={12588}, 
            state={Giza},
            country={Egypt}}

\affiliation[2]{organization={Graduate School of Information Science, University of Hyogo},
             city={Kobe},
            postcode={650-0047}, 
            country={Japan}}          
\affiliation[3]{organization={Advanced Medical Engineering Research Institute, University of Hyogo},
             city={Himeji},
            postcode={670-0836}, 
            country={Japan}}  
            

\author[1]{Ghada Khoriba}[orcid=0000-0001-7332-0759]
\ead{ghadakhoribar@nu.edu.eg}

\author[1]{Mina Yousef}[orcid=0009-0000-0316-2724]
\ead{myousef@nu.edu.eg}

\author[2,3]{Essam A. Rashed}[orcid=0000-0001-6571-9807]
\ead{rashed@gsis.u-hyogo.ac.jp}
\cormark[1]




\cortext[1]{Corresponding author}



\begin{abstract}
Medical Image Segmentation (MIS) stands as a cornerstone in medical image analysis, playing a pivotal role in precise diagnostics, treatment planning, and monitoring of various medical conditions. This paper presents a comprehensive and systematic survey of MIS methodologies, bridging the gap between traditional image processing techniques and modern deep learning approaches. The survey encompasses thresholding, edge detection, region-based segmentation, clustering algorithms, and model-based techniques while also delving into state-of-the-art deep learning architectures such as Convolutional Neural Networks (CNNs), Fully Convolutional Networks (FCNs), and the widely adopted U-Net and its variants. Moreover, integrating attention mechanisms, semi-supervised learning, generative adversarial networks (GANs), and Transformer-based models is thoroughly explored.

In addition to covering established methods, this survey highlights emerging trends, including hybrid architectures, cross-modality learning, federated and distributed learning frameworks, and active learning strategies, which aim to address challenges such as limited labeled datasets, computational complexity, and model generalizability across diverse imaging modalities. Furthermore, a specialized case study on lumbar spine segmentation is presented, offering insights into the challenges and advancements in this relatively underexplored anatomical region.

Despite significant progress in the field, critical challenges persist, including dataset bias, domain adaptation, interpretability of deep learning models, and integration into real-world clinical workflows. This survey serves as both a tutorial and a reference guide, particularly for early-career researchers, by providing a holistic understanding of the landscape of MIS and identifying promising directions for future research. Through this work, we aim to contribute to the development of more robust, efficient, and clinically applicable medical image segmentation systems.

\end{abstract}


\maketitle

\section{Introduction}
Year after year, the volume of medical data grows exponentially as patients seek clinicians' consulting services for diagnostic and therapeutic services. Various imaging modalities have historically been used for noninvasive medical imaging to provide medical doctors with information on the patient's state; these include computed tomography (CT), magnetic resonance imaging (MRI), positron emission tomography (PET), ultrasound (US), optical coherence tomography (OCT), and X-rays \citep{Hussain5164970}. The choice of imaging modality depends on the organ and pathology investigated \citep{jardim2023image}. 

In addition to being a laborious and time-consuming process, human interpretation of medical images is vulnerable to subjectivity due to different levels of experience, as well as the possible mental and physical fatigue of the clinical expert \citep{jardim2023image}, which could potentially lead to variations in the interpretation of the same medical image by two different experts. Intra and interexpert variability is further exacerbated by irregularities in the targeted structures, morphological variations, and pathological deformities between patients \citep{LI2022102360}. This problematic irregularity in human perception led to the development of computer-aided tools that facilitate image analysis, producing precise, fast, repeatable, and objective measurements. Image analysis can be further subcategorized into image enhancement, registration, classification, segmentation, detection, localization \citep{sistaninejhad2023review}, and visualization \citep{du2020medical}.

Medical Image Segmentation (MIS) is crucial for separating anatomical and pathological structures by identifying their boundaries \citep{Babu202315}. This is done by dividing the image into homogeneous (or correlated) regions that share similar characteristics. The goal is to separate anatomically distinct organs, tissues, and lesions. This process converts the image into a more meaningful representation that can be easily analyzed \citep{jardim2023image, ramesh2021review}.

Given the wide range of imaging modalities, each presenting unique challenges and requiring specific strategies, medical image segmentation (MIS) must adapt to modality-specific characteristics and underlying physics~\citep{Islam2023,shung2012principles}. Computed Tomography (CT) relies on high-contrast density differences, making threshold-based and region-growing methods effective. Magnetic Resonance Imaging (MRI), with its rich soft tissue contrast, often benefits from deep learning models that account for spatial and intensity variations. Positron Emission Tomography (PET), which provides functional imaging, is typically segmented using fusion techniques that integrate anatomical information from CT or MRI. Ultrasound (US) images are affected by speckle noise and variability, necessitating adaptive filtering and machine learning approaches. Optical Coherence Tomography (OCT), frequently used in ophthalmology, requires layer segmentation techniques using graph-based or deep learning methods. X-ray images, often challenged by overlapping structures, benefit from deep convolutional networks and attention-based models. Understanding these modality-specific differences is essential for designing robust segmentation algorithms suited to diverse clinical applications.

Two major types of MIS are commonly used and distinguished: semantic segmentation and instance segmentation \citep{wang2022medical}. Semantic segmentation involves classifying each pixel in an image into a predefined class. This means all pixels belonging to a particular class (e.g., tumor, organ) are labeled with the same identifier. Instance segmentation classifies each pixel and distinguishes between different instances of the same class. For example, it can differentiate between multiple tumors in a single image. While both semantic and instance segmentation are crucial for medical image analysis, they serve different purposes. Semantic segmentation focuses on classifying each pixel into a class, which helps identify regions of interest.
In contrast, instance segmentation goes further by distinguishing between individual instances of the same class, which is essential for detailed analysis and quantification tasks \citep{bioengineering11101034}. It can facilitate the detection of microcalcifications in mammograms and tumor volume and automatic counting of blood cells \citep{jardim2023image}. Furthermore, it can provide the necessary spatial and volumetric information to assist in quantitative analysis and the consequent diagnosis of the patient state \citep{jiao2023learning}. It is particularly essential in Radiotherapy (RT) as it can provide the needed segmentation of CT scans based on which the physician decides the dose to administer based on a computerized radiotherapy planning system (RTPS) \citep{ramesh2021review}.

In short, MIS plays a pivotal role in the region of interest (ROI) extraction, lesion quantification, and 3D reconstruction \citep{shao2023application}. However, accurate computer-aided segmentation faces multiple challenges \citep{jardim2023image, shao2023application}:

\begin{itemize}

\item Being of an interdisciplinary nature, the field of MIS requires extensive cooperation between clinicians and machine learning scientists. 

\item Noise and other acquisition-associated artifacts in medical images make it difficult to be processed than natural images. This challenge is exacerbated by discrepancies between different imaging modalities and variations within the same modality (e.g., different X-ray machines producing images with varying contrasts and noise levels).

\item Existing medical data are limited due to the difficulty and costliness of acquiring annotated medical images. Traditional MIS Approaches (discussed in Section \ref{s3}) require substantial data to process images effectively. 

\item Inherent fallibility of specific imaging modalities. For example, soft tissues and lesions are ambiguous in CT scans, and the anatomical structure of bones is not well delineated in MRI images. This challenges segmenting regions with missing edges and a lack of texture contrast. 

\item Due to variations in spatial characteristics of images, as well as in the objective behind segmentation and the nature of the original image taken, it is challenging to develop a universally applicable segmentation method that can be executed in clinical trials.

\end{itemize}

Due to these challenges, recent research has proposed a range of solutions using diverse deep learning neural network architectures. This work aims to present a comprehensive tutorial survey covering various mechanisms and paradigms related to deep learning-based medical image segmentation (MIS), from pre-training strategies to advanced models. While most existing surveys tend to emphasize specific branches of deep learning, this paper adopts a broader perspective. It provides balanced attention to traditional, current, and emerging trends, making it especially useful for junior researchers entering the field. Additionally, we conclude the paper with a case study on lumbar spine segmentation. The main contributions of this paper are as follows:

\begin{itemize} 

\item Reviews related literature and outlines this survey’s unique contributions (Section~\ref{s2}). 

\item Provides a tutorial on traditional image segmentation methods that remain relevant today, including thresholding, edge detection, region-based segmentation, clustering, model-based segmentation, and graph-based approaches (Section~\ref{s3}). 

\item Offers a chronological overview of deep learning-based MIS, from conventional convolutional neural networks to fully convolutional networks and U-Nets, which form the foundation of many contemporary MIS techniques (Section~\ref{s4}). 

\item Examines semi-supervised learning methods in MIS, covering pseudo-labeling, unsupervised regularization, prior knowledge embedding, generative adversarial networks (GANs), and contrastive learning (Section~\ref{s5}). 

\item Highlights current and emerging trends in MIS, including cascaded networks, attention mechanisms, medical transformers, neural architecture search, cross-modality segmentation, distributed learning, active learning, uncertainty quantification, and lightweight networks (Section~\ref{s6}). 

\item Presents a case study on vertebral lumbar spine segmentation, including medical context, segmentation techniques, and recent contributions utilizing U-Nets and autoencoders (Section~\ref{s7}). 

\item Discusses limitations in current research and proposes possible future research directions. 

\end{itemize}

Medical image segmentation is a cornerstone of modern medical imaging, enabling precise diagnostics and effective treatment planning. Despite notable progress, continued research is essential to overcome existing challenges and enhance the efficiency, robustness, and accuracy of segmentation techniques. This survey aims to serve as a foundational tutorial for both traditional and state-of-the-art approaches in the field, encouraging informed contributions and future advancements in medical image segmentation research.

\begin{figure*}
\centering
\includegraphics[width=0.98\linewidth]{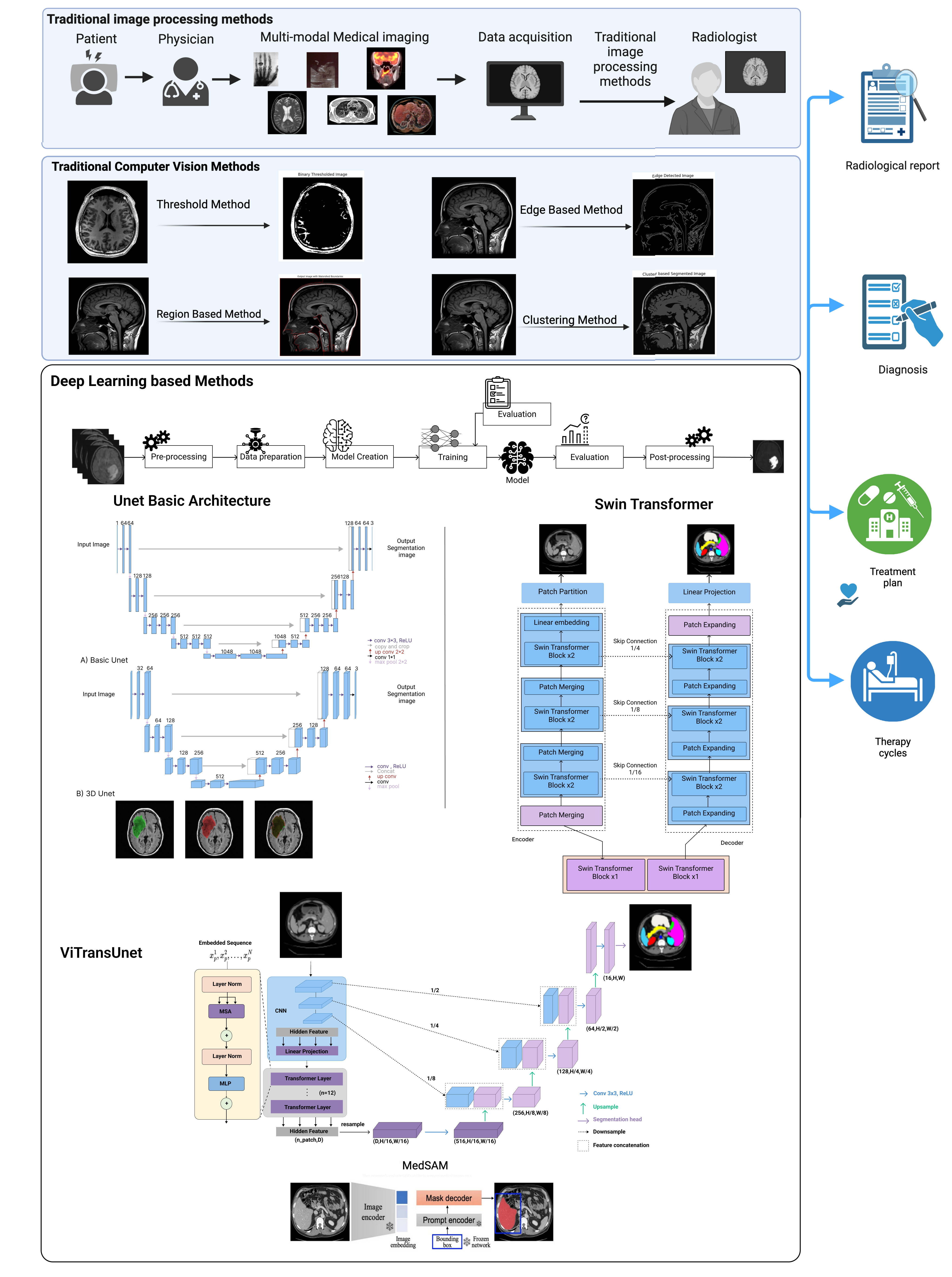}
\caption{Medical image segmentation methods summary. Traditional image processing methods are mainly data-driven approaches. Traditional computer vision methods are based on extracting spatial image features. Deep learning methods are commonly based on complex neural networks that can extract image features without prior hand-crafting.}
\label{R3_fig01}
\end{figure*}

\section{Related Works}
\label{s2}
Many survey papers have attempted to delineate various aspects of contemporary MIS research. In \citep{du2020medical}, the classification was modality-based; the authors distinguished between six imaging modalities (CT, MRI, US, X-ray, OCT, and PET scans) and surveyed the works in each. A survey of recent breakthroughs in the field of MIS was conducted in \citep{conze2023current}. This survey focuses on technical challenges and emerging research trends, such as knowledge distillation, contrastive learning, medical Transformers, prior knowledge embedding, cross-modality analysis, federated learning, and active learning. The authors in \citep{siddique2021u} primarily focused on different U-net architectures (see Section \ref{s4} for further details on the concept); they reviewed variants of U-nets and their applicability to various imaging modalities. The most recent trends and optimizations of image thresholding, a traditional segmentation technique still relevant to contemporary literature, were surveyed in \citep{jardim2023image}. The work in \citep{jiao2023learning} focuses exclusively on semi-supervised learning methods for MIS and categorizes them into pseudo-labels, unsupervised regularization, and knowledge priors. They also discussed the limitations and research directions of existing semi-supervised approaches. The state-of-the-art deep learning techniques were reviewed in \citep{sistaninejhad2023review}, and special attention was given to their application in radiology. In \citep{ramesh2021review}, contemporary optimizations to traditional image segmentation methods were reviewed, including region-based techniques, clustering techniques, edge detection, and model-based techniques; furthermore, the Lattice Boltzmann method was given special attention. The authors in \citep{wang2022medical} presented a comprehensive survey of recent trends in deep learning, including neural architecture search, graph convolutional networks, multi-modality data fusion, and medical Transformers. In \citep{shao2023application}, particular attention was given to recent works on clustering algorithms as well optimizations to U-net, and a discussion ensued about extending traditional clustering algorithms using U-net. The work in \citep{liu2021review} offers a comprehensive review of DL-based image segmentation, focusing on supervised frameworks like fully convolutional networks and U-nets and unsupervised frameworks like generative adversarial networks. They have also extensively classified the currently available medical datasets for deep learning research. 

While most of these works offer extensive and insightful overviews of current works, they often emphasize a few branches of the field over others, missing out on the inter-relatedness of different branches. As such, our survey aims to present the broad spectrum of ongoing research in the MIS field and to delineate the regions in which they overlap and those in which they complement each other  (Fig.~\ref{R3_fig01}). As such, the primary contribution of this work lies in its relative comprehensiveness, bringing together other previous works and highlighting their interrelatedness, as well as its unique distinction of being tutorial in nature, to aid researchers in the field aiming to produce novel contributions.

\section{Traditional Medical Image Segmentation Approaches}
\label{s3}
Before the rise of data-driven segmentation methods, most research focused on mathematical models and low-level image processing techniques~\citep{s10462-023-10631-z}, including thresholding, edge detection, region-based clustering algorithms, graph theory, and model-based segmentation. 

\subsection{Thresholding}

The thresholding process entails dividing the image into three categories of pixels that are either smaller than, equal to, or greater than a predetermined threshold value \citep{jardim2023image}. There are two thresholding techniques: global and local (adaptive) thresholding. 


Global thresholding views the image as a bimodal histogram, a deep valley between two distinct peaks \citep{jardim2023image}, where one peak represents the target object and the peak represents the background. Object extraction then occurs by comparing pixel values with a threshold \citep{davies2004machine}, yielding a binary image with pixel values taking either 0 (background) or 255 (object). Global thresholding techniques include the Otsu method \citep{otsu1975threshold}, the Kittler-Illingworth method \citep{kittler1986minimum}, and entropy-based global thresholding \citep{kapur1985new}. While computationally simple and fast, global thresholding only works well for images that contain objects with uniform intensity values on a contrasting background, however, it fails if the image is noisy, the contrast is low or the background intensity varies significantly across the image \citep{rogowska2000overview}. 


Local thresholding involves either splitting the image into sub-images and calculating thresholds for each or examining intensity in the neighborhood of each pixel and then determining the threshold based on intensity distribution \citep{jardim2023image}. It is more computationally expensive than global thresholding but works well in significant background variations or when the object is small. 

\subsection{Edge detection}

Relying on sudden changes in color or intensity, edge detection is one of the most fundamental image segmentation methods. It uses derivatives to determine edge pixels in an image first, then connects them to form boundaries \citep{schober2021edge}. First derivatives are usually calculated using the Roberts, Prewitt, and Sobel operators, whereas second derivative operators include Laplace and Kirch \citep{shao2023application}. The Canny edge detector applies non-maximum suppression followed by hysteresis thresholding, achieving higher accuracy and fewer broken edges \citep{archa2018segmentation}. 

\subsection{Region-based segmentation}

The region-based segmentation method uses the local spatial information of the image to combine pixel similarity into segmentation results \citep{hua2018automatic}. This is achieved using region growing, region split and merge, and the Watershed approach \citep{ramesh2021review}. 


In region growing, a seed point is first selected in the image. Then, the region expands by accepting neighboring pixels that share specific criteria with the seed, including pixel intensity, color, or spatial proximity \citep{ranjbarzadeh2023me}. 


Region split and merge process entails first dividing the image into large regions, using grid partitioning or quadtree decomposition \citep{ramesh2021review}. Each region is recursively subdivided into smaller regions until each small region has pixels sharing similar qualities based on predefined criteria. Finally, to avoid over-segmentation, the small regions are merged based on similar attributes or neighborhoods. It is computationally expensive for complex images; however, the hierarchical nature of the process aids multi-level analysis of the image. 


Particularly useful for segmenting adjacent regions, the watershed approach relies on intensity gradients to delineate boundaries between objects, treating pixel intensities as topographical features \citep{ramesh2021review}. The idea is that low-intensity pixels are viewed as valleys and high-intensity pixels are viewed as peaks \citep{pavlidis2012algorithms}, water flows from the peaks to the valleys, and where water from two different peaks meets at the same valley, a boundary is created. This method is particularly effective in boundary preservation and works well with complex images without prior information. However, it is computationally expensive and vulnerable to over-segmentation if markers (peaks) are not accurately chosen. 

\subsection{Clustering}

Clustering is the process of dividing pixels into homogeneous regions sharing similar values. It is an unsupervised, efficient, and self-adaptive process \citep{jiang2020novel}. It can be sub-categorized into hard clustering and soft (fuzzy) clustering. 


The traditional method of hard clustering is the K-means clustering algorithm. The idea is to choose K centroids and then assign each data point in the image to a specific centroid based on Euclidean distance, which segments the image into K clusters. By taking the average value of the coordinates of all points in a particular cluster, a new cluster centroid can be specified and repeated until no significant change occurs between two successive iterations \citep{shao2023application}. 
Improvements to K-means include crowd-based (FGO) optimization \citep{abhiraj2021enhanced}, median filtering, Sobel edge detection and morphological operations \citep{agrawal2018segmentation}, two-stage fuzzy K-means \citep{husein2018implementation}, combining K-means with discrete wavelet transform \citep{kumar2021improved} and Darwinian particle swarm optimization \citep{mehidi2019improved}. 


Contrary to hard clustering, in soft clustering algorithms, data points can belong to multiple clusters simultaneously, offering a more nuanced representation of cluster points. It is more robust to noise than K means and allows for greater flexibility since the degree of overlap (fuzziness) can be controlled. The traditional soft clustering algorithm is the fuzzy C-means algorithm, improvements to which include fuzzy local intensity clustering, fuzzy clustering based on spatial information, and new fuzzy clustering algorithms, which are reviewed in detail in \citep{shao2023application}. 


A graph-based representation of an image attributes each pixel to a node on the graph, and the connections between adjacent pixels are represented as edges. The weight of the edge is a measure of similarity between neighboring pixels in terms of gray level, color, or texture. The target then becomes to divide the graph into sub-graphs that share the maximum possible feature similarity \citep{shao2023application}. Two possible techniques are graph cut \citep{chen2018survey} and grab cut \citep{rother2004grabcut}, where the former is a one-time energy minimization algorithm, and the latter is a more interactive iterative process.


In model-based segmentation, continuous curves express the target edges in an image using parametric and geometric contour models \citep{fang2024image}. 


Active Shape Models (ASM) is a statistical shape model constructed from a training set of different positions and orientations of the same object to capture its variations and constraints \citep{cootes1992active}. Then, an iterative process commences with the intention of finding the best model to fit the statistical shape and the new image. 

Active Appearance Models (AAM) use two statistical models: shape and appearance (including texture). This yields a more robust object representation at the expense of greater computational complexity \citep{cootes2001active}. 

\begin{figure}
\centering
\includegraphics[width=0.48\textwidth]{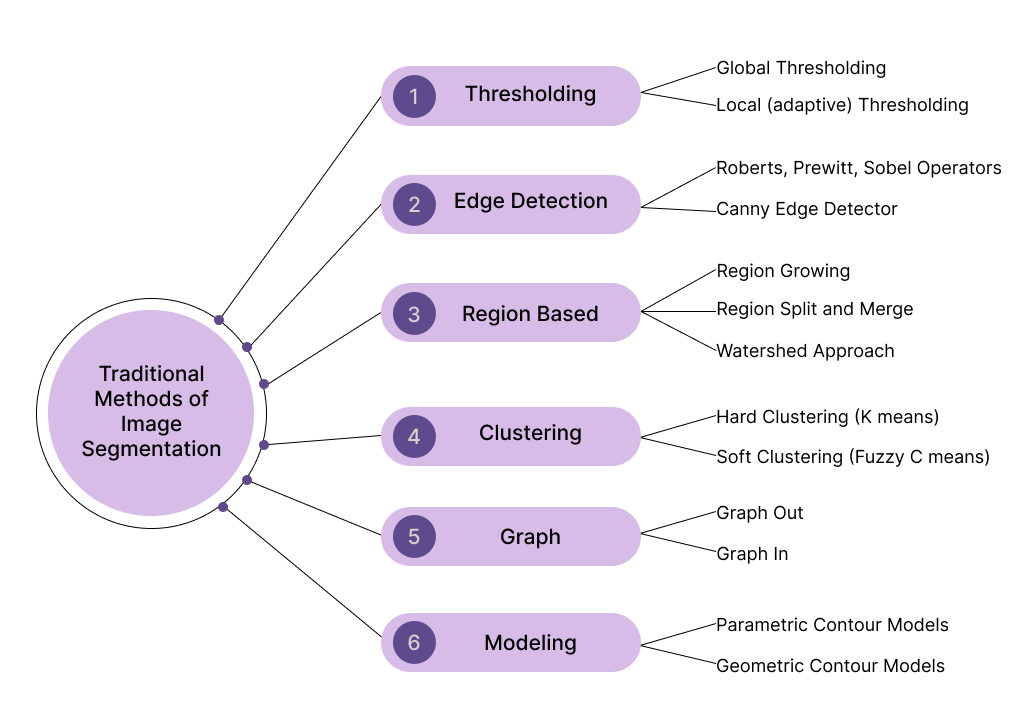}
\caption{Traditional methods of image segmentation.}
\label{R3_fig02} 
\end{figure}

Rather than using mathematical functions to describe the contours, geometric contours comprise points and vertices, yielding greater accuracy at the expense of reduced flexibility compared to parametric contours  \citep{osher1988fronts}. Geometric contours outperform parametric contours in their ability to handle topological changes in curves and their insensitivity to initial positions. Figure \ref{R3_fig02} summarizes traditional methods of image segmentation.

\subsection{Strengths and limitations}
Traditional medical image segmentation approaches provide a foundational framework for extracting meaningful information from medical images, offering both notable strengths and limitations. One of their primary advantages lies in their simplicity and interoperability, supported by well-established mathematical principles. Techniques such as thresholding and edge detection are computationally efficient, easy to implement, and require minimal data, making them suitable for real-time applications and resource-constrained environments. Region-based segmentation methods leverage spatial coherence, enhancing segmentation accuracy when objects exhibit clear intensity homogeneity. Clustering techniques, particularly K-means and fuzzy C-means, offer flexible, self-adaptive solutions that require limited prior knowledge.
Despite these advantages, traditional methods face several critical limitations. Global thresholding often performs poorly under non-uniform illumination or low contrast, leading to inadequate object-background separation. Edge detection is highly sensitive to noise and artifacts, frequently producing fragmented or incomplete boundaries that necessitate post-processing. Region-based methods can be computationally intensive and prone to over-segmentation, especially in noisy or highly textured images. Clustering techniques struggle with determining the optimal number of clusters and may be influenced by initialization biases. Moreover, traditional approaches generally lack adaptability to complex, high-dimensional medical images, reducing their effectiveness in addressing anatomical variability, pathological complexity, and multimodal datasets. These limitations have contributed to the growing adoption of machine learning and deep learning-based methods, which are capable of learning high-level representations and adapting to the complexity of medical imaging challenges.

Table \ref{tab:comparison} summarizes a comparison of traditional medical image processing techniques (such as thresholding and edge detection) with deep learning methods (such as CNNs and U-Nets) in terms of advantages, limitations, and use cases, including metrics like accuracy, speed, and robustness to noise.

\begin{table*}
\centering
\caption{Comparison between traditional and deep learning methods.}
\label{tab:comparison}
\resizebox{\textwidth}{!}{%

\begin{tabular}{|l|p{6cm}|p{6cm}|}
\hline
\textbf{Aspect} & \textbf{Traditional Methods } & \textbf{Deep Learning Methods} \\ \hline

\textbf{Advantages} & 
\begin{itemize}
    \item Simple and easy to implement \citep{Shah202395253,Babu202315}
    \item Low computational cost \citep{Shah202395253}
    \item Effective for well-defined edges and contrasts \citep{Shah202395253}
\end{itemize}
& 
\begin{itemize}
    \item High accuracy and generalizability \citep{Shah202395253,Babu202315,Chen2023,Kavyasri20242970}
    \item Robust feature extraction \citep{Shah202395253,Babu202315,Chen2023}
    \item Adaptable to various medical imaging tasks \citep{Babu202315,Chen2023,Kavyasri20242970}
\end{itemize} \\ \hline

\textbf{Limitations} & 
\begin{itemize}
    \item Low classification accuracy \citep{Shah202395253}
    \item Poor robustness to noise and variations \citep{Shah202395253,Babu202315}
    \item Limited to simple segmentation tasks \citep{Shah202395253,Babu202315}
\end{itemize}
& 
\begin{itemize}
    \item High computational complexity \citep{Kavyasri20242970}
    \item Requires large annotated datasets \citep{Juneja2024,Arumaiththurai2021185}
    \item Potential overfitting and generalizability issues \citep{Shah202395253,Kavyasri20242970}
\end{itemize} \\ \hline

\textbf{Use cases} & 
\begin{itemize}
    \item Basic segmentation tasks \citep{Shah202395253,Babu202315}
    \item Initial preprocessing steps \citep{Sneha2023809}
\end{itemize}
& 
\begin{itemize}
    \item Complex segmentation tasks (e.g., tumors, organs) \cite{Babu202315,Kavyasri20242970,Chassagnon2020}
    \item Disease detection and classification \citep{Cai2024,An2020,Özkaraca2023}
\end{itemize} \\ \hline

\textbf{Accuracy} & Generally lower accuracy \citep{Shah202395253,Babu202315} & Higher accuracy \citep{An2020} \\ \hline

\textbf{Speed} & Faster for simple tasks \citep{Shah202395253} & Slower due to high computational demands \citep{Kavyasri20242970} \\ \hline

\textbf{Robustness to noise} & Poor robustness \citep{Shah202395253,Babu202315} & Better robustness with advanced architectures \citep{Shah202395253,Kavyasri20242970,Lou2021} \\ \hline
\end{tabular}%
}
\end{table*}


\section{Deep learning based Medical Image Segmentation}
\label{s4} 

While traditional MIS methods are not contemporarily irrelevant, their frequency in recent literature is diminishing compared to data-driven segmentation techniques based on deep learning. The inherent capabilities of DL allow it to learn about abstract features of data at different levels, enabling the detection of image morphology and texture patterns \citep{du2020medical}. 
Good at observing hidden patterns in images \citep{sistaninejhad2023review}, DL has facilitated robust image segmentation across various diseases, anatomies, and imaging modalities \citep{Lou2021,Shah202395253,Kavyasri20242970} by enabling quantitative analysis and 3D visualization of medical images \citep{du2020medical}. Therapeutic planning, follow-up, prognostic, dosimetric, and radionics applications have been concretely affected by the development of DL models \citep{nafchi2025radiomics}. However, while inputs and outputs to a DL network are precise, the behavior of hidden layers is ambiguous, and their functionality cannot be easily replicated or understood, which is why DL has not yet been applied to any large-scale real-world medical trials despite exhibiting tremendous promise \citep{siddique2021u}. 

\subsection{Convolutional Neural Networks}

Deep learning simulates the human brain's learning process using neural networks \citep{liu2021review}, enabling it to extract features from large-scale data without human supervision. The Convolution Neural Network (CNN) is the classic DL model used in image processing \citep{gu2018recent}. Many accomplishments have been achieved in image feature extraction, pattern recognition, and classification owing to CNNs \citep{liu2021review}. Proposed in 2012, AlexNet \citep{krizhevsky2017imagenet} was distinguished for its success in image classification. CNN have become a popular and effective tool for this purpose due to their ability to learn and extract features from images. However, limitations of CNNs include their overreliance on geometric priors, rendering it difficult to fully capture the intrinsic relationships between different objects using extracted local features \citep{wang2022medical}. Graph CNNs were proposed as a powerful and intuitive alternative for non-Euclidean spaces, which enabled the exploitation of intrinsic relationships to extract otherwise invisible connections between objects. 

\begin{figure}
\centering
\includegraphics[width=0.48\textwidth]{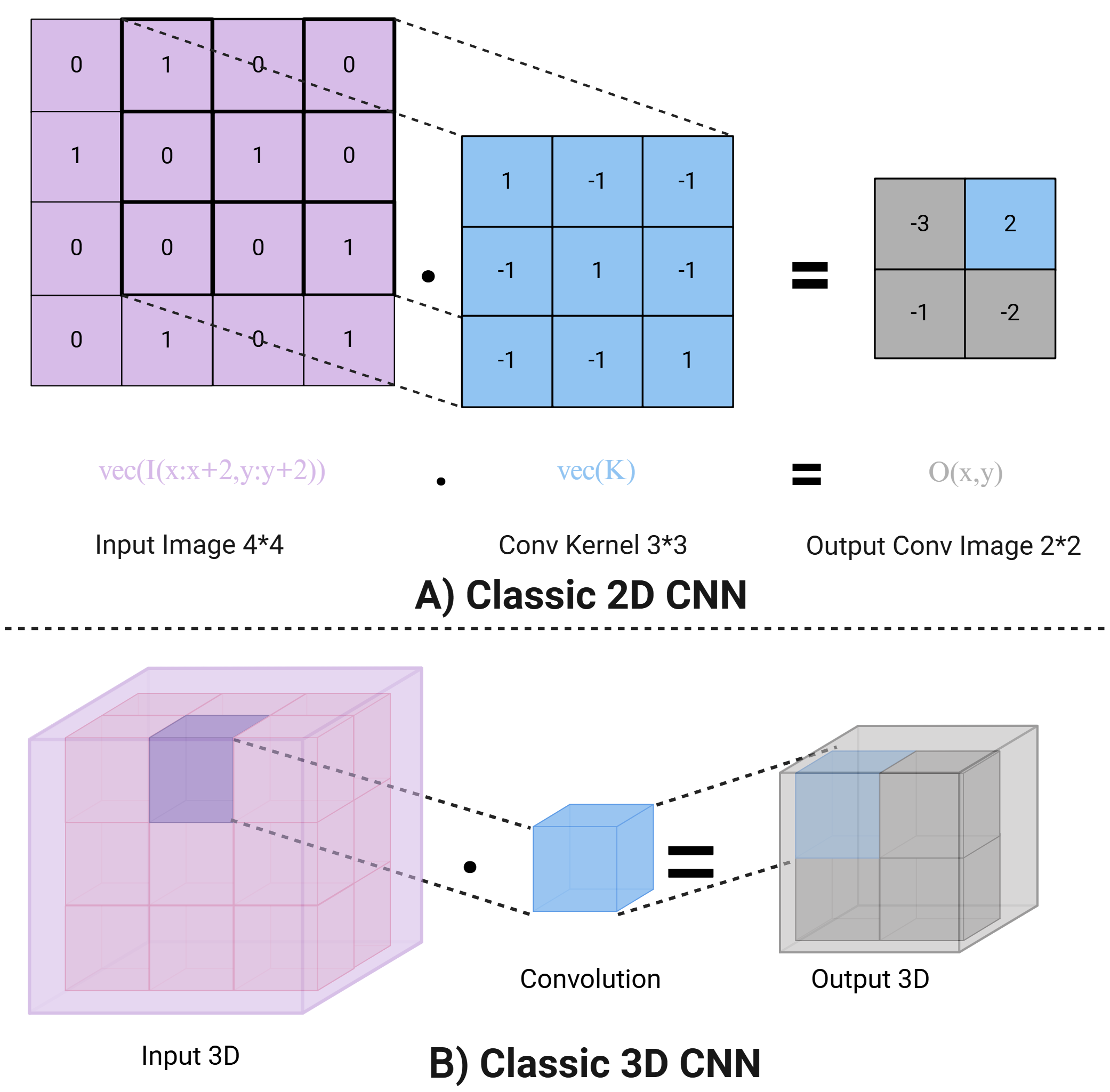}
\caption{Classic (a) 2D and (b) 3D CNN and the convolution operation \citep{liu2021review}.}
\label{R3_fig03}
\end{figure}

The CNN comprises multiple hidden layers sandwiched between an input layer and an output layer and is responsible for various tasks such as convolution, pooling, and activation \citep{liu2021review}. The input layer is connected to the input image, comprising several neurons matching the pixels of the image. The second layer is a convolution layer, which performs feature extraction on the input data to yield a feature map; this is determined by parameter setup in the convolution kernel. The next layer is a pooling layer, which is responsible for filtering and selecting feature maps to simplify the computational complexity of the network. Then, a fully connected layer connects all neurons in the previous layer to yield an output, which is finally sent to the classifier \citep{liu2021review}. Examples of classic 2D CNNs are ResNet \citep{he2016deep} and Visual Geometry Group (VGG) \citep{simonyan2014very}. The classic 2D CNN is shown in Fig. \ref{R3_fig03} (a).

Medical images are essentially three-dimensional, even though clinicians tend to analyze 2D slices of these images. However, to effectively investigate these images, the convolution kernel must be 3D as well \citep{qiu2017learning}, which enables the extraction of more powerful volumetric representations and spatial considerations. An example of 3D CNN is shown in Fig.~\ref{R3_fig03} (b).

\begin{figure*}
\centering
\includegraphics[width=0.8\textwidth]{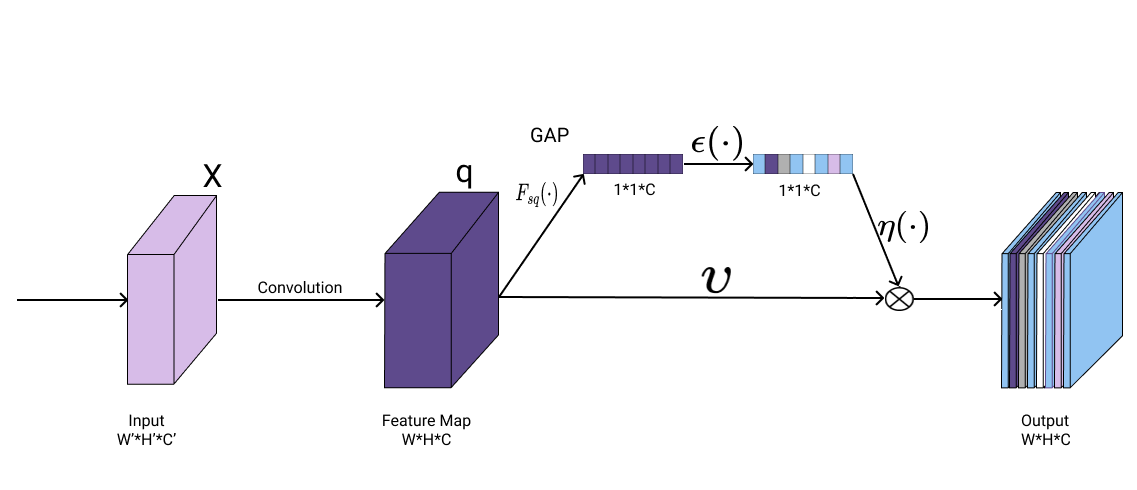}
\caption{Flowchart diagram of squeeze and excitation block \citep{hu2018squeeze}. GAP stands for global average pooling.}
\label{R3_fig04}
\end{figure*}

While increasing the number of network layers or the network depth can yield better results, the network is vulnerable to other problems like overfitting and vanishing gradients \citep{liu2021review}. To solve this problem, GoogleNet \citep{szegedy2015going} proposed an inception structure that increases the depth and width of the network while maintaining or reducing the number of parameters, which is achieved using multiple convolution kernels of different sizes and adding pooling. ResNet \citep{he2016deep} also solved the problems associated with network depth using residual blocks, where each module consists of several consecutive layers and a shortcut that connects the input and output layers of the module before ReLU activation. Finally, squeeze and excitation blocks \citep{rundo2019use} (Fig. \ref{R3_fig04}) improve the expressive ability of the network. 

The encoder-decoder architecture, enhanced with CNNs and Transformers, has proven effective for image segmentation tasks. Innovations in boundary detection, attention mechanisms, and multi-scale feature fusion continue to improve the accuracy and efficiency of these models, making them suitable for various applications, particularly in medical imaging. SegNet \citep{badrinarayanan2017segnet} is a deep CNN architecture designed for semantic image segmentation. It employs an encoder-decoder structure to perform pixel-wise classification, making it suitable for various applications such as urban scene understanding, medical image processing, and autonomous driving. These include (PSP Net) \citep{zhao2017pyramid}, in which a pyramid pool module and a pyramid scene parsing network were proposed to aggregate context information of different regions into global context information in a pyramid-like scheme. In \citep{ren2015faster}, an instance segmentation framework (Mask R-CNN) is proposed.

\subsection{Fully Convolutional Networks (FCN)}

\begin{figure*}
\centering
\includegraphics[width=0.9\textwidth]{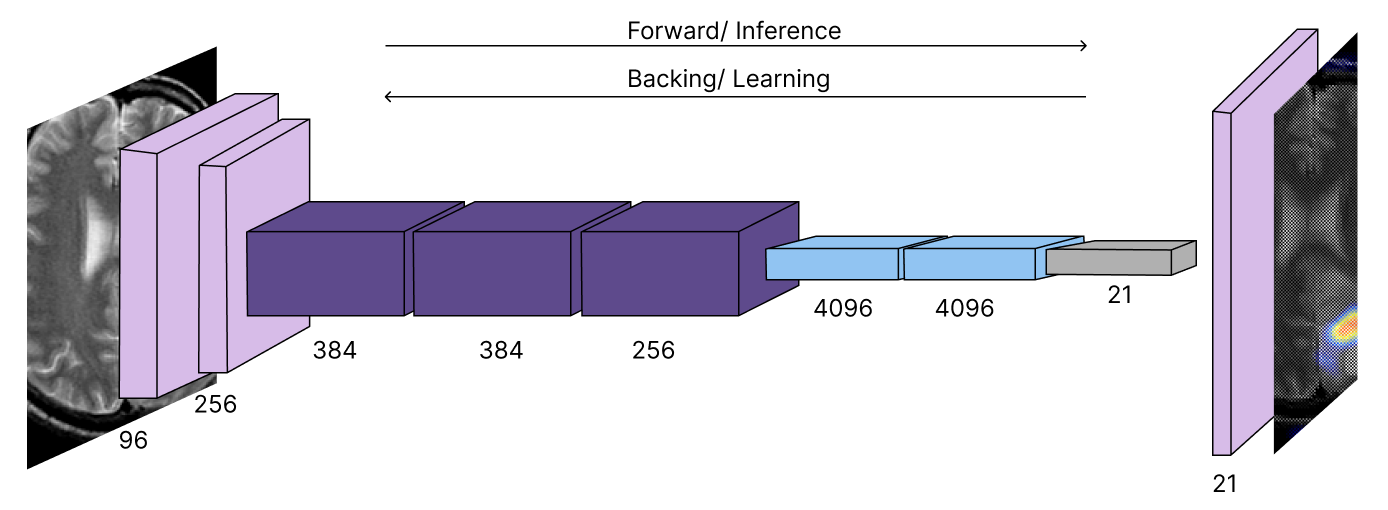}
\caption{Structure of FCN used for brain tumor segmentation in MRI~\citep{long2015fully}.}
\label{R3_fig05}
\end{figure*}

The traditional CNN structure comprises convolutional layers followed by fully connected layers, which means that the network's final output is one-dimensional \citep{liu2021review}; this makes it suitable for image classification and object detection tasks. However, in image segmentation, we require pixel-wise prediction rather than categorizing the image as a whole \citep{long2015fully}. Fully Convolutional Networks (FCNs) improve traditional CNNs by removing the fully connected layers from their architecture and replacing them with convolutional layers instead; this enables the attainment of a 2D feature map of each pixel. Furthermore, FCNs can accept any image size at the input and use deconvolution layers to upsample the last convolution layer's feature map and restore the input image's size. 
However, upsampling in the traditional FCN yields fuzzy results insensitive to the image's details. Thus, improvements to the conventional FCN were proposed, including DeepLab v1 \citep{chen2014semantic}, which is inspired by VGG16 with atrous convolutions and Conditional Random Field (CRF), DeepLab v2 \citep{chen2017deeplab}, which is inspired by spatial pyramid pooling but introduces atrous spatial pyramid pooling (ASPP) with a parallel convolutional sampling of holes at different sampling rates on a given input, and DeepLab v3 which introduced a cascaded atrous convolution module. Furthermore, a 2.5D approach to FCN was implemented in \citep{zhou2017deep}, which implements three FCNs for each 2D profile but works better for larger organs \citep{liu2021review}. In \citep{zhou2017focal}, it was proposed to apply focal loss on FCN to reduce the number of false positives in medical images due to class imbalance. The FCN schematic is shown in Fig. \ref{R3_fig05}.
FCNs for 3D MIS have evolved significantly, with various innovative approaches addressing the challenges of volumetric data, computational complexity, and limited training samples. Models like AdaEn-Net, CMV convs, and UNETR represent the forefront of this field, achieving state-of-the-art results across multiple medical imaging tasks \citep{BaldeonCalisto202076, Li2021571, Hatamizadeh20221748}.

\subsection{U-Net}

\begin{figure*}
\centering
\includegraphics[width=0.82\textwidth]{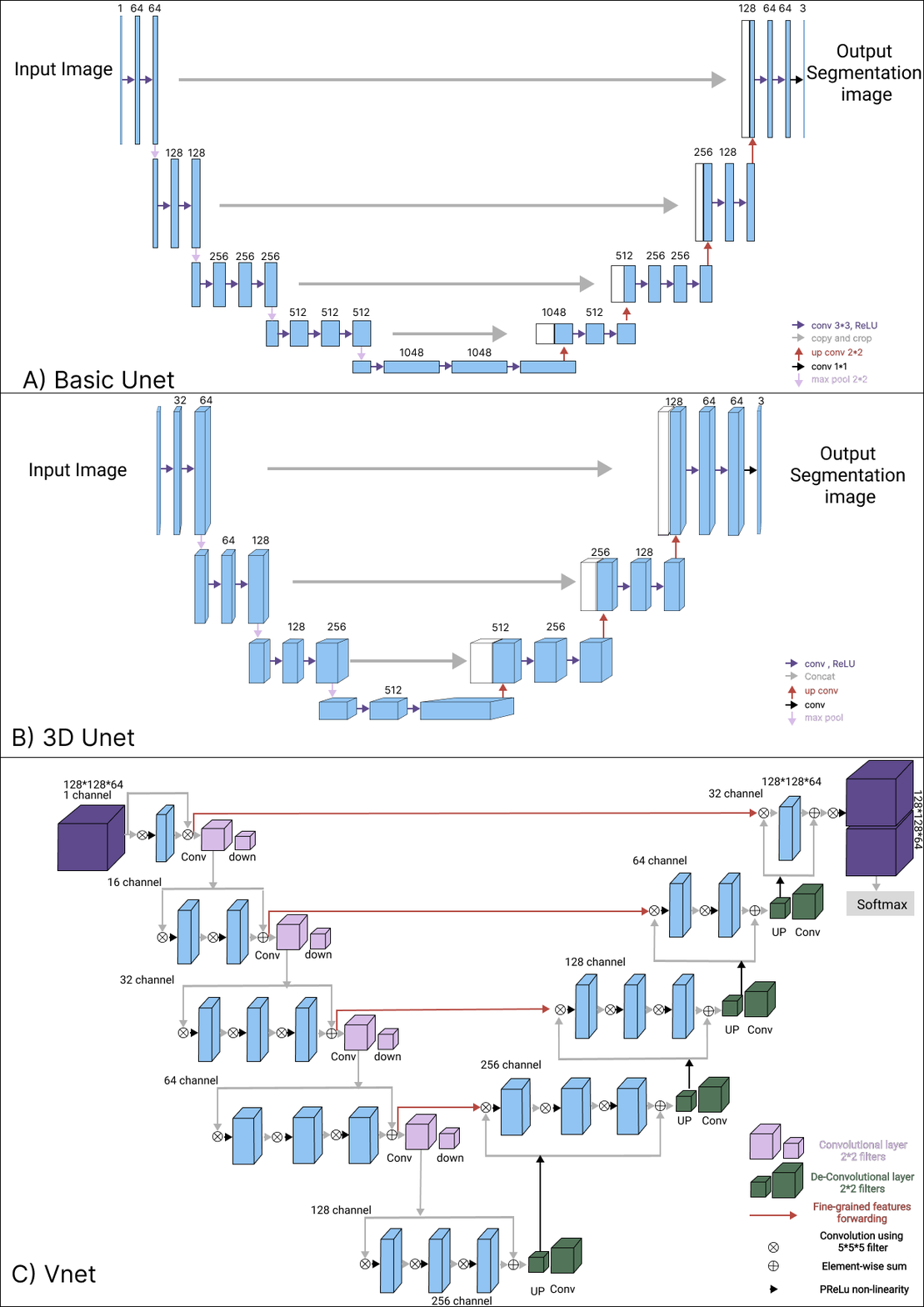}
\caption{A) Basic U-net architecture. Blue boxes represent the feature map, and gray boxes represent the cropped feature maps at each layer \citep{siddique2021u}. B) 3D Unet architecture. C) V-net architecture.}
\label{R3_fig06}
\end{figure*}

In 2015, inspired by FCNs, Ronneberger et al. \citep{ronneberger2015u} designed a U-Net network for MIS. It is composed of a U-shaped channel similar in structure to the SegNet encoder-decoder architecture, where the encoder employs successive pooling layers to reduce spatial dimension, and the decoder progressively recovers object resolution using upsampling. The contracting path encoder typically consists of sequential 3$\times$3 convolutional layers followed by batch normalization (BN) and rectified linear unit (ReLU) activation. Spatial size is reduced using 2$\times$2 max-pooling layers. The decoder is built symmetrically concerning the encoder, the only exception being that max-pooling layers are replaced with transpose convolution, bilinear interpolation, or any other upsampling operation. A final layer with SoftMax activation achieves pixel-wise segmentation at the original resolution. Skip connections improve localization accuracy and convergence speed by concatenating features between contracting and expanding paths \citep{conze2023current}. U-Net's suitability for MIS stems from its ability to combine low-level information (for accuracy) with high-level information (to extract complex features), therefore propagating contextual information along the network, which allows it to segment objects in an area using context from a larger overlapping area \citep{siddique2021u}. The U-Net architecture is shown in Fig. \ref{R3_fig06} (a). U-net received rapacious attention from the MIS community, prompting many improvements and developments that gave credence to the base U-net as their starting point. They are summarized next.

%
Three-dimensional U-Net and its variants have significantly advanced 3D MIS, offering high accuracy and efficiency in handling complex medical imaging tasks.
The 3D U-Net architecture offers significant advantages, making it highly effective for medical imaging tasks. It is specifically designed to handle volumetric data, which is crucial for processing medical imaging modalities like MRI and CT scans that provide 3D images \citep{Almajalid20191725, Jia2019221}. The architecture supports end-to-end training, optimizing the segmentation process using objective functions such as the Dice coefficient to address the imbalance between foreground and background voxels \citep{Jia2019221, Chai2021208}. Additionally, 3D U-Net excels in feature extraction of multiple adjacent slices as input to capture more contextual information, which improves segmentation performance \citep{Almajalid20191725}.

Despite its advantages, 3D U-Net faces challenges, particularly its computational complexity and the large amount of data required. To mitigate these issues, various modifications and optimizations have been introduced. For instance, the Self-Excited Compressed Dilated Convolution (SECDC) module reduces computational load while maintaining high segmentation accuracy by combining normal and dilated convolutions \citep{Yang2021}. The Convolutional Block Attention Module (CBAM) enhances feature representation and segmentation accuracy through attention mechanisms \citep{Guo2024881}. Furthermore, contour loss refines segmentation by incorporating additional distance information to improve output precision \citep{Jia2019221}.

Several variants and modifications of the 3D U-Net have been developed to enhance performance. The Y-Net variant employs dilated convolutions to capture features at multiple scales, improving the segmentation of small anatomical structures \citep{Kemassi2022428}. The Half-UNet simplifies the encoder and decoder components, reducing computational requirements while maintaining performance \citep{Lu2022}. The Enhanced 3D U-Net also integrates spatial attention mechanisms to improve feature extraction and model robustness \citep{Dong2023}.

These advancements have enabled 3D U-Net models to achieve remarkable results in various applications. For brain tumor segmentation, enhanced models have demonstrated high Dice coefficients and robust performance \citep{Yang2021, Guo2024881}. In knee MRI segmentation, adjacent slices' modifications have improved accuracy and performance \citep{Almajalid20191725}. Similarly, dual 3D U-Net structures have precisely segmented the left atrium, with high sensitivity and specificity \citep{Jia2019221}.

This combination of innovations and applications highlights the versatility and impact of 3D U-Net in advancing MIS. Fig. \ref{R3_fig06} shows the architecture of U-Net (A), 3D U-Net (B), and V-Net (C).





\begin{figure*}
\centering
\includegraphics[width=0.9\textwidth]{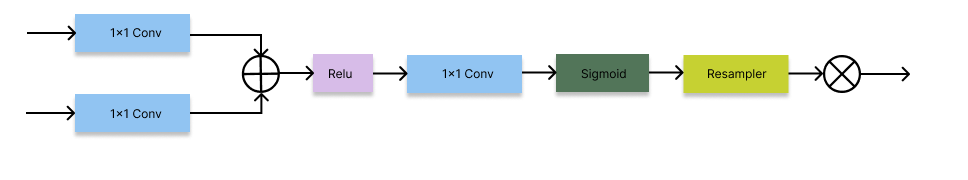}
\caption{Attention block in Unet \citep{oktay2018attention}.}
\label{R3_fig07}
\end{figure*}

Attention U-nets \citep{oktay2018attention,schlemper2019attention} exploit attention gates to trim features irrelevant to the current task, focusing specifically on essential objects in an image \citep{siddique2021u}. Each layer in the expansive path has an attention gate through which the corresponding features from the contracting path must pass before concatenating with the upsampled features. Improved segmentation occurs due to localized classification information rather than global information without a significant increase in computational complexity. Figure~\ref{R3_fig07} shows an additive attention gate. 

\begin{figure*}
\centering
\includegraphics[width=0.9\textwidth]{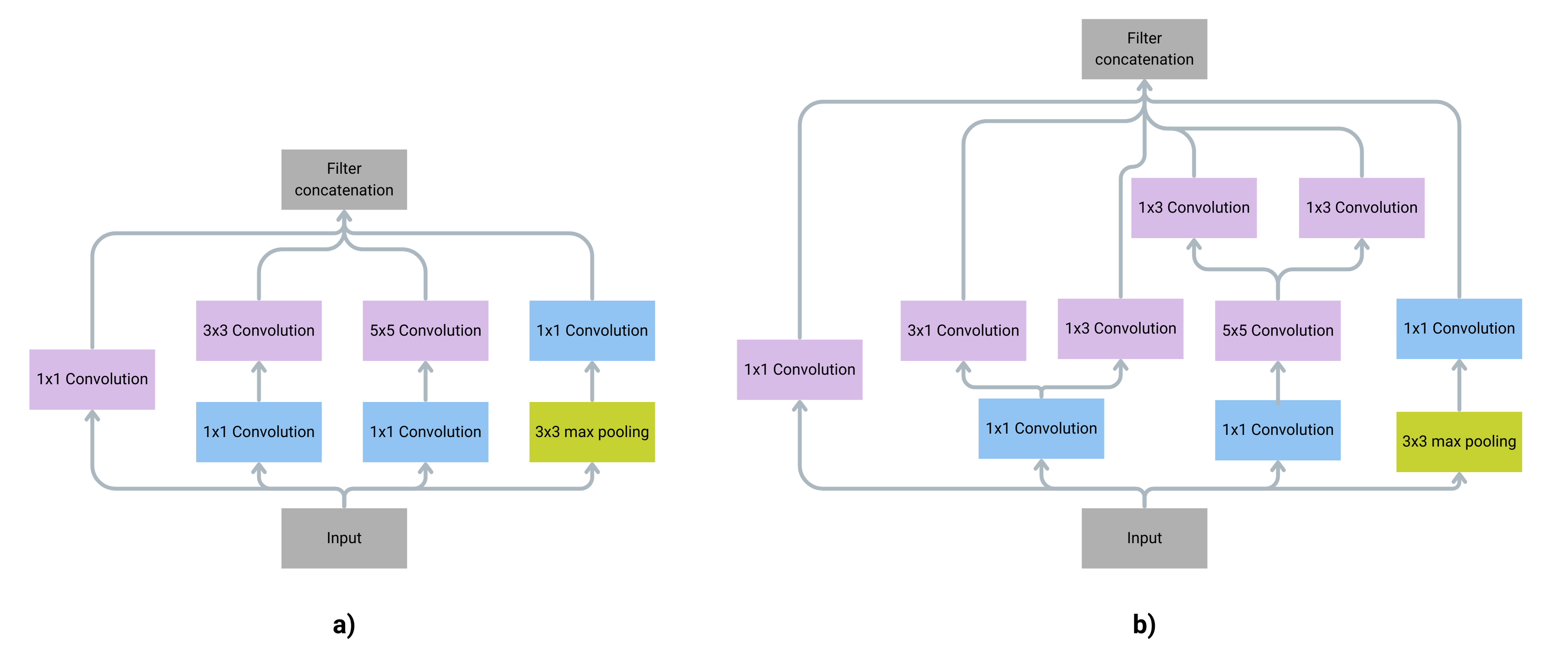}
\caption{The original inception module used in GoogleNet (b) Improved inception block with factorized filters, it yields an equivalent effect to (a) but at less computational power \citep{siddique2021u}.}
\label{R3_fig08}
\end{figure*}

It is often the case that segmentation requires looking at images with considerable variations in shapes and sizes; this is achievable using inception networks. Inception networks can effectively analyze images with different salient regions because they use multiple-sized filters on the same layer in the network \citep{siddique2021u}. GoogleNet \citep{szegedy2015going} proposed the original inception network. This was followed by multiple improvements that achieved equivalent performance at substantial computational cost reduction, simply by replacing 5$\times$5 convolutions with two successive 3$\times$3 convolutions. Figure~\ref{R3_fig08} presents two configurations of the inception module.

\begin{figure*}
\centering
\includegraphics[width=0.9\textwidth]{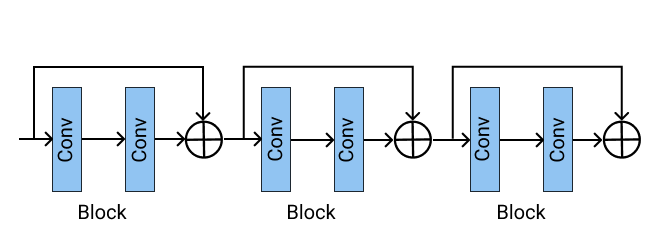}
\caption{Successive ResNet blocks with skip connections.}
\label{R3_fig09}
\end{figure*}

The ResNet inspires residual U-nets \citep{he2016deep} architecture, whose primary motivation was to overcome difficulties associated with training deep networks. The basic idea is that increasing the depth of the network yields faster convergence but at the expense of performance degradation due to the loss of feature identities caused by diminishing gradients \citep{siddique2021u}. ResNet tackles this problem by utilizing skip connections, which add the feature map of one layer to another layer deeper in the network, hence preserving the feature map. Figure~\ref{R3_fig09} displays the basic residual block architecture. 

\begin{figure}
\centering
\includegraphics[width=0.48\textwidth]{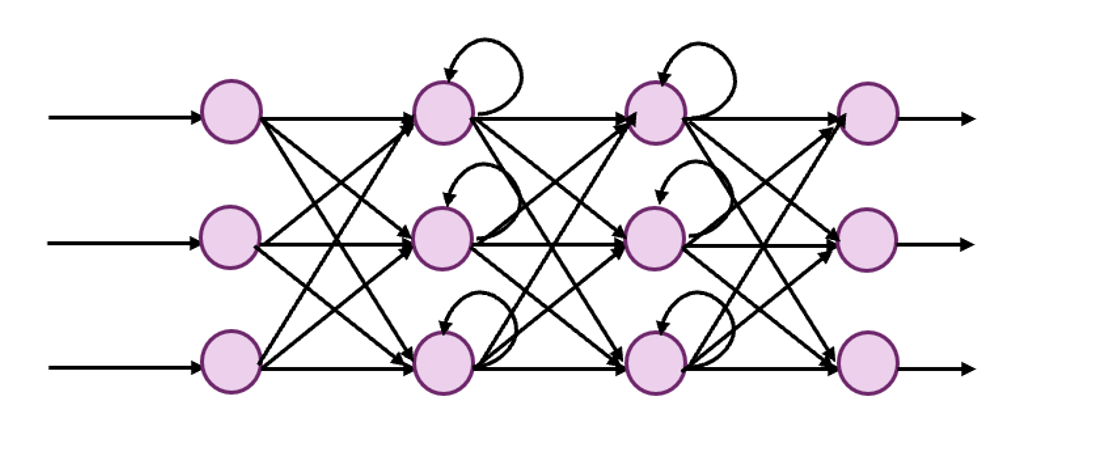}
\caption{RNN architecture.}
\label{R3_fig10}
\end{figure}

Initially designed to analyze sequential data, recurrent neural networks (RNNs) employ feedback loops (or recurrent connections) so that a node's output is affected by the previous output \citep{siddique2021u}. The recurrent U-net employs recurrent convolutional neural networks (RCNNs) \citep{liang2015recurrent}, which allows units to use context from adjacent units to update their feature maps. The architecture is shown in Fig.~\ref{R3_fig10}. 

\begin{figure}
\centering
\includegraphics[width=0.48\textwidth]{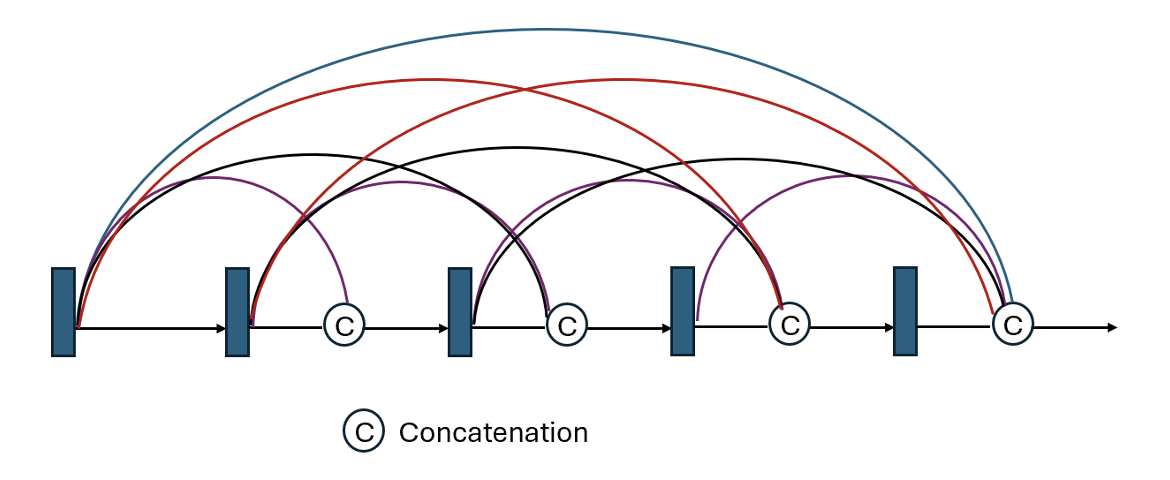}
\caption{DenseNet architecture.}
\label{R3_fig11}
\end{figure}

DenseNet \citep{huang2017densely} is an extension of ResNet to better resolve the problem of vanishing gradients. This is achieved by complementing every layer in a block with feature maps from all preceding layers and combining feature maps into tensors using channel-wise concatenations rather than element-wise additions, as in ResNet. This significantly promotes gradient propagation, and each layer can have fewer channels as information is better preserved across layers \citep{siddique2021u}. This is shown in Fig.~\ref{R3_fig11}. 

\begin{figure}
\centering
\includegraphics[width=0.48\textwidth]{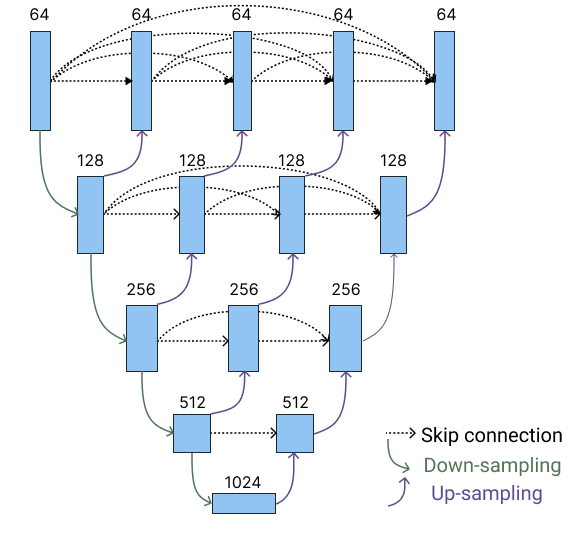}
\caption{Unet$^{++}$ architecture \citep{siddique2021u}. An improved version of basic Unet (Fig.~\ref{R3_fig06} A) with additional skip connections that enable better capture fine-grained features and reduce the semantic gap between encoder and decoder feature maps.}
\label{R3_fig12}
\end{figure}

Inspired by DenseNet, Unet$^{++}$ (Fig.~\ref{R3_fig12}) employs a dense network of skip connections between the contracting and expanding paths, aiding in the propagation of more semantic information between the two paths. It is beyond the scope of this paper to detail the vast number of different U-net architectures proposed in the literature, and we restricted ourselves to delineating the fundamental building blocks upon which other U-net architectures are based. To name a few, ensemble U-nets (or cascaded U-nets) are architectures combining two or more U-nets, where the first U-net performs a high-level segmentation and then each successive U-net performs segmentation on smaller objects \citep{siddique2021u}. Parallel U-nets were proposed in \citep{abd2019tpuar} and their results were aggregated for improved accuracy, a 2.5D U-net employs 3 U-net networks running in parallel on three different 2D projections of a 3D image yielding a 3D segmentation map at a smaller computational cost than a 3D U-net. 3D attention context Unet was proposed in \citep{kang2020acu} with spatial attention blocks to aid in 3D context guidance \citep{shao2023application}. Multi-scale nested U-net (MSN U-net) \citep{fan2021msn} is conceptually similar to Dense U-net in that the semantic gap is alleviated; however, they achieve this using a multi-scale context fusion block that combines the top and bottom layers \citep{shao2023application}. RU-Net \citep{leclerc2019ru} combines the base U-net architecture with a multi-level boundary detection network using an image segmentation algorithm that employs a multi-layer boundary perception self-attention mechanism. Finally, no new Unet (nnU-net) \citep{isensee2021nnu} is a standardization attempt designed to tackle data set diversity by condensing and automating key decisions for designing a successful pipeline for any given dataset. nnUnet surpasses most existing approaches and achieves state-of-the-art performance while automatically configuring the strategies of pre-processing, training, inference, and post-processing to any arbitrary dataset \citep{jiao2023learning}.  nnU-Net automatically configures itself for any new task, including preprocessing, network architecture, training, and post-processing, without requiring manual intervention \citep{isensee2021nnu, Sabrowsky-Hirsch202089}. One limitation of nnU-Net is the lack of uncertainty measures, which can be problematic in heterogeneous datasets. Recent advancements have introduced methods to estimate uncertainty without altering the original architecture, enhancing segmentation accuracy and quality control \citep{Zhao2022535}. 

Several U-Net variations have been proposed to address challenges in MIS, such as low contrast, class imbalance, and variability in anatomical structures. As discussed above, a widely used variants are Attention U-Net, ResUNet, and nnU-Net. These models demonstrate different features. Attention U-Net is known to enhances segmentation near complex boundaries and small structures which is  useful in organ delineation and it requires slightly larger memory size in comparison with vanilla U-Net. ResUNet is known to improves gradient flow and paralytically performs well in noisy datasets and for multimodality applications. However, it requires additional memory resourses to accommodate with residual blocks. nnU-Net is considered a leading architecture that reported state-of-the-art segmentation accuracy performans in several applications. Nevertheless, auto-configuration leads to complex pipelines that requires relatively high memory and computational powers.

Among their many superior attributes, U-nets accept images of arbitrary size, solve problems related to shadow and overlap \citep{du2020medical}, and create highly detailed segmentation maps using samples with minimal annotation \citep{siddique2021u}. This is achieved through the random elastic deformation of data \citep{ronneberger2015u} as well as the use of context-based learning. Another advantage is its incredible modularity and mutability, facilitating an avalanche of new and improved versions that improve its structure without changing its essential foundational structure. However, compressing the network model without reducing stability remains an open problem \citep{shao2023application}, and they are still dependent on high-quality labeled datasets. The semi-supervised learning approach discussed in the following section is an alternative philosophy to reduce this dependence. 
Table \ref{comparison_key_architectures} compares the MIS key architectures. 

\begin{table*}
\centering
\caption{Comparison of key network architectures.}
\label{comparison_key_architectures}
\begin{tabular}{|p{2cm}|p{3cm}|p{4cm}|p{4cm}|}
\hline
\textbf{Architecture} & \textbf{Key Features}               & \textbf{Advantages}                                                                 & \textbf{Limitations}                                                                 \\ \hline
U-Net                 & Encoder-decoder with skip connections & High accuracy, widely used, effective for 2D and 3D images                          & May struggle with complex tasks without modifications                                \\ \hline
U-Net++               & Nested, dense skip pathways          & Reduces semantic gap, improves performance                                         & Increased complexity and computational cost                                          \\ \hline
Attention U-Net       & Attention mechanisms                & Focuses on relevant regions, better for complex tasks                              & May require more computational resources                                             \\ \hline
3D U-Net              & 3D convolutional layers             & Suitable for volumetric data                                                       & Higher computational and memory requirements                                         \\ \hline
ViT                   & Transformer-based                   & Powerful performance, captures global context                                      & Limited generalizability, requires adaptation for different tasks                   \\ \hline
Hybrid Models         & Combines CNNs and Transformers      & Balances local and global features, efficient                                      & Complexity in design and implementation                                              \\ \hline
Active Contour        & Deformable models                   & Maintains smooth boundaries, good for complex structures                           & May require manual tuning and initialization                                         \\ \hline
SAM                   & Zero-shot learning, generalizes across modalities & Versatile, adaptable                                                               & Initial performance may be lower without fine-tuning                                 \\ \hline
\end{tabular}

\end{table*}

\subsection{Challenges}

Deep learning has significantly advanced the field of MIS, offering automated and accurate delineation of anatomical and pathological structures. However, despite its success, the application of deep learning methods in MIS presents several notable challenges that limit their broader clinical adoption. One of the primary limitations is the dependency on large, high-quality annotated datasets. Medical image annotation is both time-consuming and resource-intensive, often requiring expert radiologists or clinicians. While some open source online resources are now available (REF), it still limited in terms of number of images and variety of clinical applications. The complexity increases when dealing with rare diseases or subtle pathologies, where inter-observer variability and limited sample availability exacerbate the challenge. Moreover, data privacy regulations and institutional barriers further restrict data sharing, making it difficult to compile diverse, representative training sets.

Another key issue is domain shift. Deep learning models trained on data from a specific institution or scanner often struggle to generalize to new environments. Differences in imaging protocols, scanner vendors, patient populations, and acquisition settings can lead to significant drops in performance when the model is applied outside of its original context. Without robust domain adaptation techniques, such models may fail in real-world clinical scenarios where consistency and reliability are critical.

Interpretability is also a pressing concern. Deep learning models, especially those based on convolutional neural networks and Transformers, are often viewed as black boxes. In clinical applications, understanding why a model made a certain segmentation decision is essential for trust and accountability. Although recent efforts in explainable AI provide some insights through attention maps or activation visualizations, these methods are still limited in terms of clinical transparency and interpretability.

Furthermore, many segmentation tasks suffer from class imbalance. Structures of interest, such as tumors or lesions, often occupy only a small portion of the image. This imbalance can bias models toward the background or majority classes, resulting in missed detections or reduced sensitivity for clinically important findings. Coupled with the high computational demands of training and deploying deep models, these issues can be prohibitive, especially in resource-limited clinical settings.

Finally, integrating deep learning models into clinical workflows involves overcoming regulatory, infrastructural, and evaluation challenges. Obtaining regulatory approval (e.g., FDA or CE certification), ensuring data privacy compliance (e.g., HIPAA, GDPR), and aligning model outputs with clinically meaningful metrics are non-trivial tasks. Traditional performance metrics such as the Dice coefficient or Intersection over Union (IoU) may not fully capture the clinical relevance of segmentation results, underscoring the need for more standardized and clinically validated evaluation protocols.

\section{Semi-Supervised Learning in Medical Image Segmentation}
\label{s5}

Despite the considerable success of DL techniques in general, and U-nets in particular, in realizing accurate image segmentation, the impracticality of obtaining large-scale carefully labeled datasets remains problematic \citep{jiao2023learning}. To ease the burden of manual labeling, considerable efforts have been dedicated to annotation-efficient DL techniques, including data augmentation, conditional generative adversarial networks, contrastive learning, and others, all succinctly described as semi-supervised learning techniques. In semi-supervised settings, we aim at building a model of comparable performance to fully supervised models but with a relatively limited amount of labeled data and a vast amount of unlabeled data. This section covers the various techniques and strategies for achieving semi-supervised medical image segmentation (SSMIS). 

\subsection{Pseudo Labels}

\begin{figure*}
\centering
\includegraphics[width=0.9\textwidth]{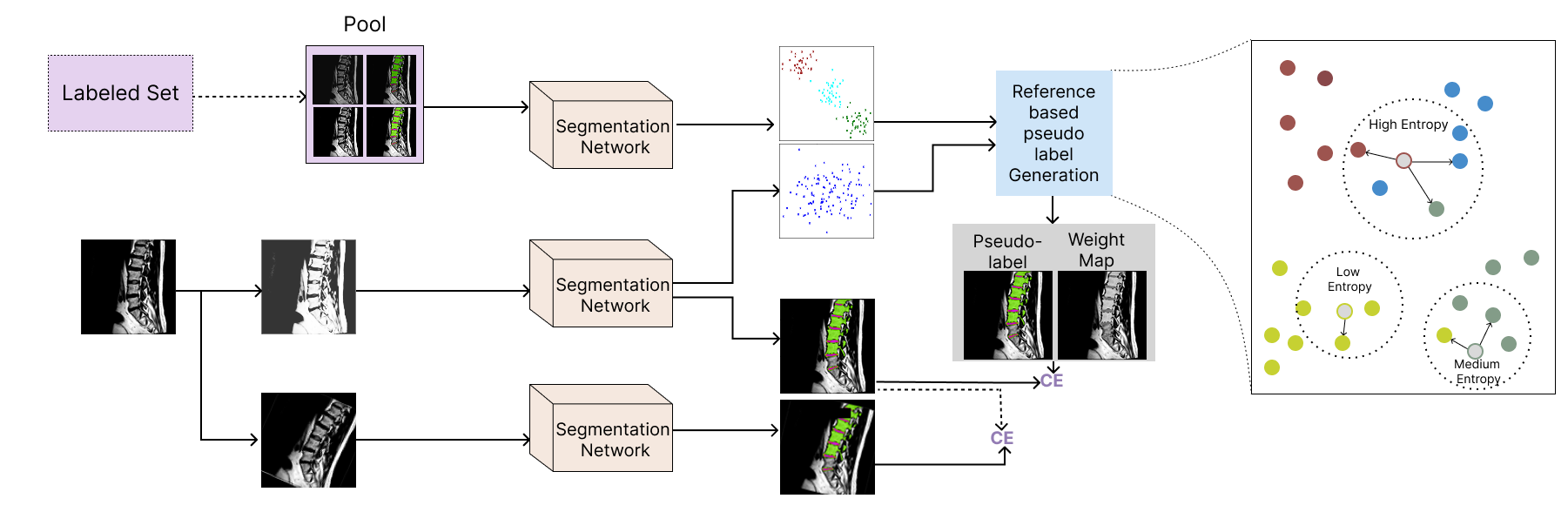}
\caption{Pseudo-label generation model \citep{seibold2022reference}.}
\label{R3_fig13}
\end{figure*}

Pseudo-labeling is a popular semi-supervised learning strategy in which pseudo annotations are generated for unlabeled data and iteratively used to improve the segmentation model. Initially, a model is trained on a small amount of labeled data and then applied to the unlabeled set to produce pseudo-segmentation masks. These pseudo-labeled samples are then combined with the original labeled data to retrain and refine the model~\citep{jiao2023learning}. This process is repeated over multiple iterations to progressively enhance model performance. A key challenge in pseudo-labeling is handling noisy predictions, as inaccurate pseudo-labels can hinder training convergence~\citep{lee2013pseudo}. Various methods have been proposed to mitigate this, differing in strategies for label quality assurance and model initialization.

Online pseudo-labeling involves using high-confidence predictions during training as temporary labels. For example,~\cite{jiao2023learning} applied this approach to the Synapse multi-organ segmentation dataset, demonstrating that using a confidence threshold improved label precision. While annotation efficiency was significantly improved—reducing the need for manual labeling by over 60\%, the quality of pseudo-labels was still contingent on threshold calibration and model stability.

Offline pseudo-labeling can be achieved through label propagation, with methods such as:

(a) Prototype learning~\citep{han2022effective}: This approach computes distances between feature vectors of unlabeled images and class prototypes. Using morphological post-processing, high-quality pseudo labels are generated. \cite{han2022effective} used the Left Atrium (LA) dataset and reported that pseudo-labeling helped reduce expert annotation time by nearly 50\% while maintaining strong segmentation accuracy.

(b) Nearest neighbor matching~\citep{wang2021neighbor}: Here, pseudo labels are assigned based on embedding similarity to neighboring labeled instances. This method was validated on the NIH pancreas CT dataset, showing that even with only 20\% labeled data, performance approached that of fully supervised methods.

Figure~\ref{R3_fig13} illustrates a typical pseudo-label generation framework. While pseudo-labeling methods offer substantial annotation savings, their effectiveness heavily depends on model initialization, data distribution, and robustness to noisy predictions.

\subsection{Unsupervised Regularization}

This approach attempts to incorporate unlabeled data into the training process with an unsupervised loss function \citep{jiao2023learning}. Different choices of the unsupervised loss function and regularization term lead to various models.

\begin{figure}
\centering
\includegraphics[width=0.48\textwidth]{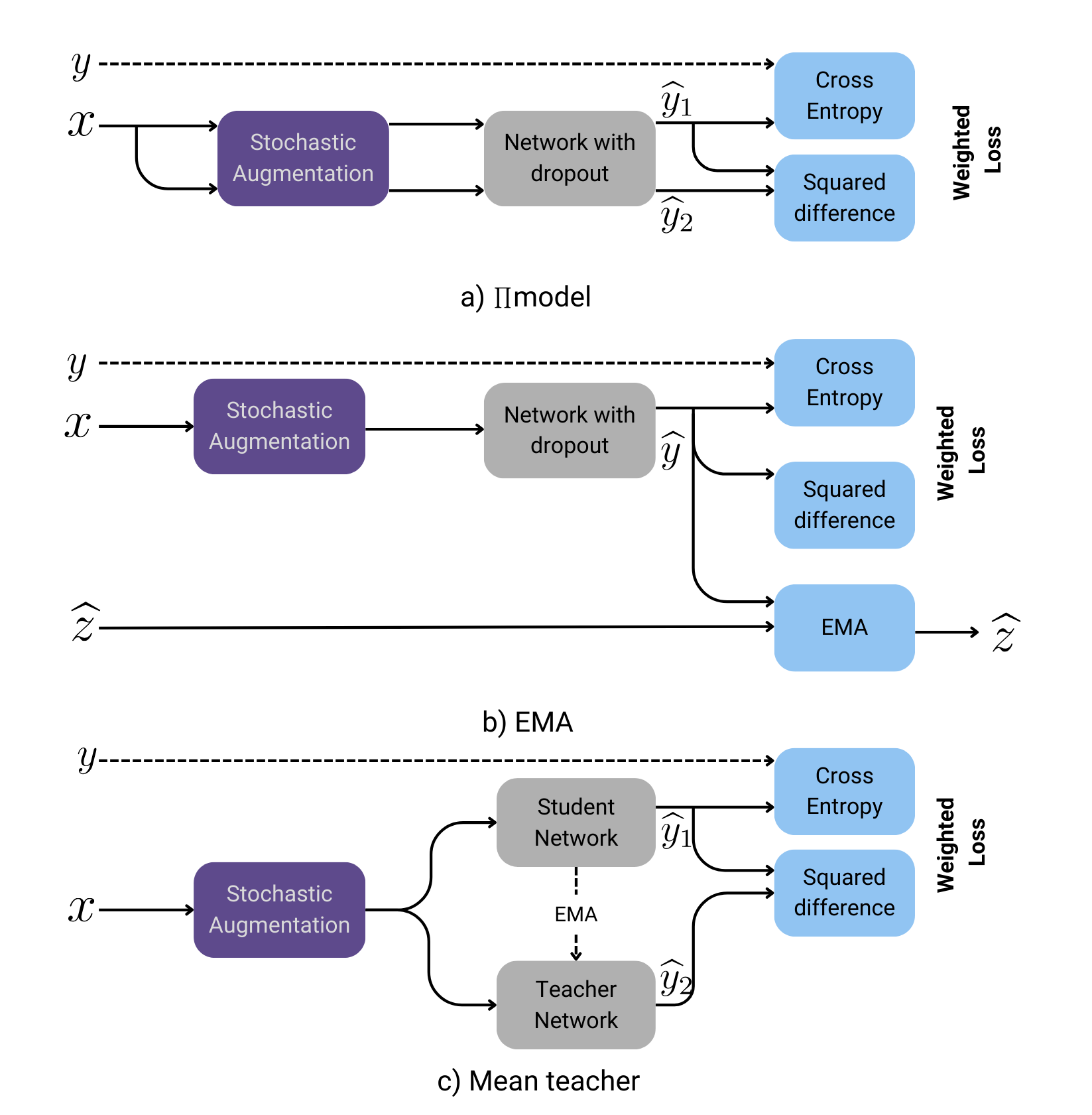}
\caption{Architectures of consistency learning. (a) $\Pi$ model, (b) EMA, (c) Mean teacher \citep{jiao2023learning}.}
\label{R3_fig14}
\end{figure}

\begin{figure}
\centering
\includegraphics[width=0.48\textwidth]{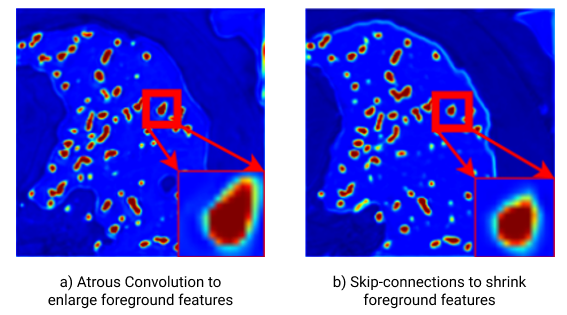}
\caption{Perturbations used in consistency learning. a) Atrous convolutions to enlarge foreground features \citep{luo2016understanding}, b) Skip connections to shrink foreground features \citep{he2016deep}.}
\label{R3_fig15}
\end{figure}

Consistency learning enforces an invariance of predictions under different perturbations and pushes the decision boundary to low-density regions \citep{jiao2023learning}, based on the assumption that perturbations would not change the model's output. The consistency between two objects is determined using similarity measures like Kullback-Leibler divergence, mean square error, or Jensen-Shannon divergence \citep{jiao2023learning}. Consistency Learning suffers from sensitivity to noise, dependence on the choice of parameters, and dependence on the choice of perturbations. There are different architectures to achieve consistent learning: 
\begin{itemize}
    \item[(a)] $\Pi$ model \citep{sajjadi2016regularization}: two random augmentations of a sample are created for both labeled and unlabeled data. The model expects consistency in the output of the same unlabeled sample propagating twice under different perturbations. 
    \item[(b)] EMA \citep{laine2016temporal}: Exponential moving average predictions are used as consistency targets for unlabeled data. However, maintaining EMA during the training is a heavy burden.
    \item[(c)] Mean Teacher Architecture: Alleviates the burden of maintaining EMA by utilizing teacher and student networks where the consistency of predictions from perturbed inputs between teacher and student is enforced. Figure~\ref{R3_fig14} shows the different architectures, and Fig.~\ref{R3_fig15} shows different perturbations; for further detail on different perturbation types, refer to \citep{jiao2023learning}. 
\end{itemize}

\begin{figure*}
\centering
\includegraphics[width=0.9\textwidth]{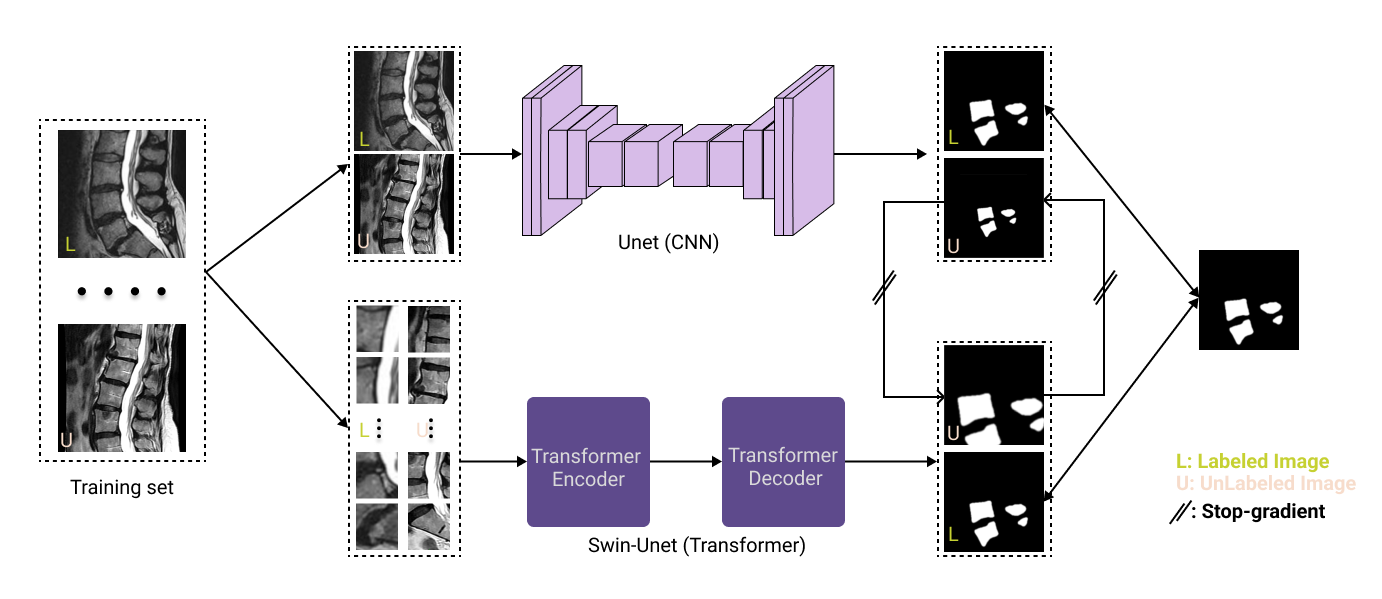}
\caption{A co-training framework \citep{luo2022semi}.}
\label{R3_fig16}
\end{figure*}

Unsupervised regularization with a co-training framework assumes two or more different views of each datum and that each view has sufficient information to generate independent predictions \citep{blum1998combining}. First, it learns a separate segmentation model for each view on labeled data; then, it gradually adds model predictions on unlabeled data to the training set to continue training. One view is assumed redundant to other views; the models are encouraged to have consistent predictions on all views \citep{jiao2023learning}. It diverges from self-training models in that pseudo labels are added from one view to the training set and then used to supervise other views. It differs from consistency training in that all models in co-training undergo gradient-descent-based updates. This approach is limited by the assumption that each view is sufficient and independent enough to generate its predictions, the risk of overfitting if the two models are too similar, and sensitivity to noisy pseudo labels. Figure~\ref{R3_fig16} shows an example of co-training architecture. 

\subsection{Prior knowledge embedding}

\begin{figure*}
\centering
\includegraphics[width=0.8\textwidth]{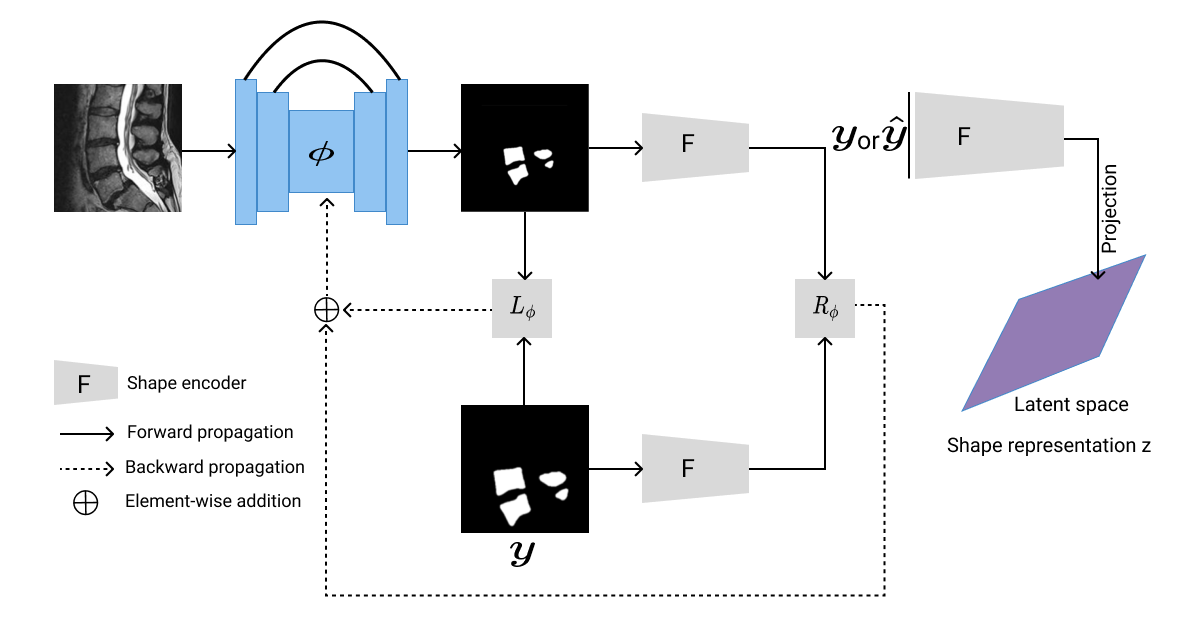}
\caption{Autoencoder architecture \citep{conze2023current}.}
\label{R3_fig17}
\end{figure*}

Different types of information can be included as prior knowledge of DL frameworks, including shape constraints, topology specifications, edge polarity, or inter-region adjacency rules \citep{conze2023current}. In \citep{oktay2018attention, boutillon2022multi}, the authors used autoencoders (AE) to demonstrate the learning of anatomical shape variations from medical images. Autoencoders follow an encoder-decoder architecture, where the encoder maps the input to a low-dimensional feature space (smaller than the input dimension), and the decoder reconstructs the image from the feature space. A typical autoencoder architecture is shown in Fig.~\ref{R3_fig17}. Priors are not restricted to shape only, in fact, texture, topology and size might be more meaningful priors to incorporate into a DL network for increased robustness \citep{ravishankar2017learning}. Another example of prior knowledge embedding was featured in \citep{oktay2017anatomically}, where the authors designed anatomically constrained neural networks (ACNN) for MIS tasks. Limitations of knowledge priors include overfitting (if prior knowledge is too specific to the training data) and non-differentiable models (if knowledge priors are too complex, like region connectivity, convexity, and symmetry). 

\subsection{Generative Adversarial Networks (GAN)}

\begin{figure}
\centering
\includegraphics[width=0.48\textwidth]{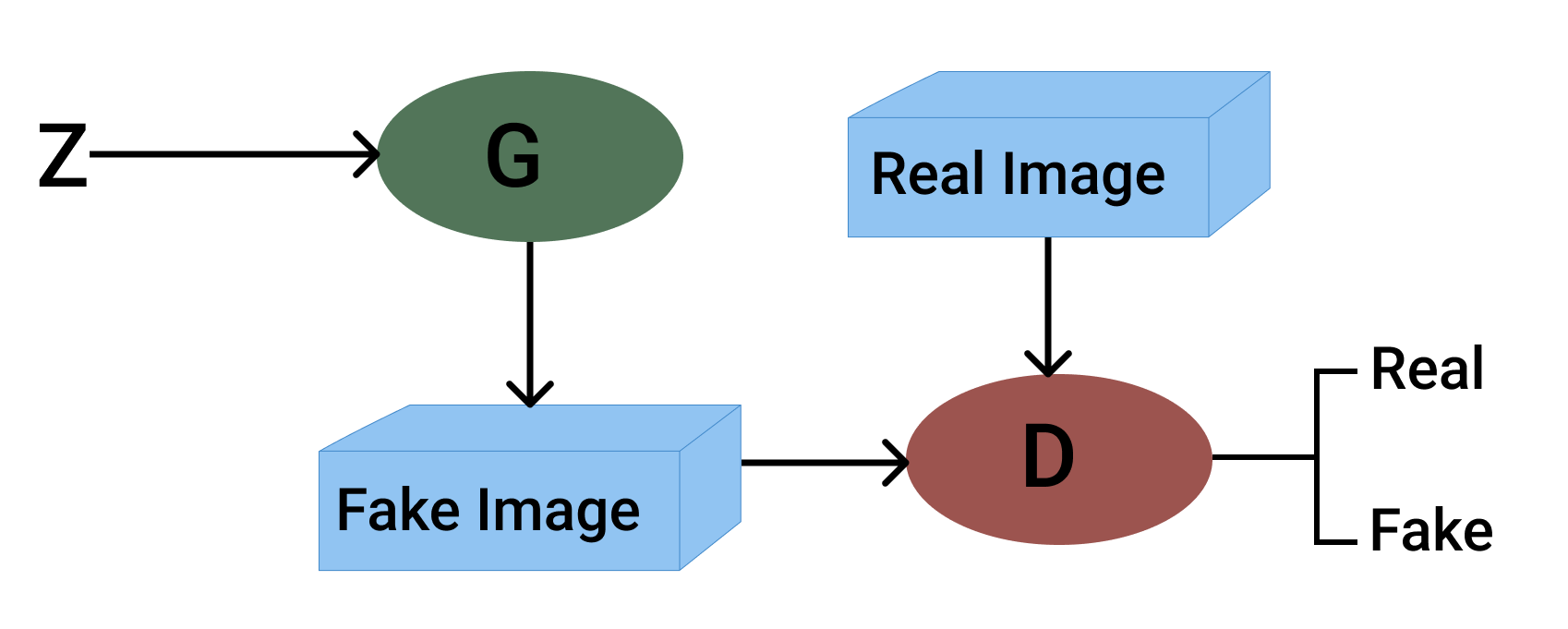}
\caption{Generative Adversarial Network basic architecture.}
\label{R3_fig18}
\end{figure}

\cite{goodfellow2014generative} introduced Generative Adversarial Networks (GANs), which consist of two neural networks, the generator and the discriminator, engaged in a competitive game. The generator creates synthetic data from random noise, while the discriminator attempts to distinguish real from fake data~\citep{liu2021review}. This adversarial process leads to progressively more realistic outputs from the generator (see Fig.~\ref{R3_fig18}).

The application of GANs to semantic segmentation was pioneered by~\cite{luc2016semantic}, who utilized a CNN-based generator to synthesize segmentation masks that closely resemble the ground truth. While initially developed using the Cityscapes dataset, this approach showed that even with a limited number of labeled examples, the model could learn realistic structure, thus improving annotation efficiency by reducing dependence on pixel-wise annotations.

Segmentation Adversarial Network (SegAN)~\citep{xue2018segan} incorporated a U-Net generator to handle medical image segmentation, where adversarial learning was used to refine outputs by minimizing the L1 loss. Evaluated on the ISBI 2012 EM segmentation dataset, SegAN achieved comparable performance to fully supervised methods using only ~50\% of the labeled data. Structure Correcting Adversarial Network (SCAN)~\citep{dai2018scan} extended this framework by using fully convolutional networks (FCNs) for both generator and discriminator. The discriminator imposed anatomical plausibility by enforcing structural consistency based on human physiology. Applied to the NIH Pancreas CT dataset, SCAN demonstrated superior boundary accuracy while reducing manual annotation effort by approximately 40\%. Projective Adversarial Network (PAN)~\citep{khosravan2019pan} introduced 3D context through 2D projections, avoiding the computational burden of 3D segmentation. The dual-discriminator design addressed global and local inconsistencies. PAN was tested on the PROMISE12 prostate MRI dataset, achieving competitive results while using only partial annotations during training.

Beyond segmentation, GANs have been employed for privacy-preserving synthetic data generation. In AsynDGAN \citep{chang2020synthetic}, a distributed framework was introduced where a central generator collaborated with multiple site-specific discriminators. Applied to the BraTS brain tumor dataset, AsynDGAN enabled multi-institutional model training without sharing patient data, effectively eliminating the need for direct annotation at centralized locations.

\begin{figure*}
\centering
\includegraphics[width=0.9\textwidth]{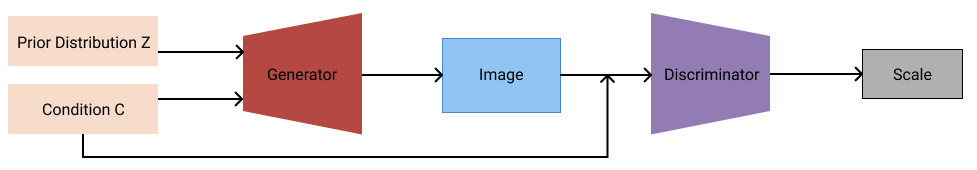}
\caption{A general sketch of cGAN architecture \citep{mirza2014conditional}.}
\label{R3_fig19}
\end{figure*}

Conditional GANs (cGANs)~\citep{bayramoglu2017towards} offer another direction, where the generation process is guided by a condition (e.g., a class label or segmentation map). Singh et al.\citep{singh2020breast} used cGANs for breast tumor segmentation with limited labeled samples, reporting strong segmentation performance while relying on only 20\% of the annotated dataset. Similarly,~\cite{wang2022medical} showed how cGAN-based augmentation could reduce annotation needs by generating diverse and anatomically consistent samples for rare pathologies. A more complex setup was presented in~\citep{guibas2017synthetic}, where a cascaded architecture of a GAN and cGAN generated label maps and corresponding synthetic images. The model was validated on synthetic datasets and later applied to real anatomical structures, showing potential for use in annotation bootstrapping workflows. A general illustration of cGAN is shown in Fig.~\ref{R3_fig19}.

Although many advanced GAN architectures exist, detailing all is beyond the scope of this review. We direct the reader to~\citep{zhao2018craniomaxillofacial, mondal2018few, zhang2017deep} for additional promising implementations in 3D medical image segmentation (MIS). In general, GAN-based frameworks significantly improve annotation efficiency, particularly in settings with limited ground truth availability or strict data privacy requirements.

\subsection{Contrastive Learning}

Contrastive Learning (CL) is a technique used to improve the quality of visual representations by contrasting semantically similar and dissimilar pairs of samples. This method has shown significant promise in medical image segmentation tasks, including vertebra segmentation, by enhancing the ability to distinguish between anatomical structures without relying heavily on labeled data. ARCO~\citep{You2023} is a semi-supervised contrastive learning framework that uses variance-reduction techniques to improve pixel/voxel-level segmentation tasks with limited labels. Another example is the OBCL~\citep{Fan2025} Own-background Contrastive Learning framework that effectively incorporates background pixels into CL to handle imbalanced data distributions.

\begin{figure*}
\centering
\includegraphics[width=0.9\textwidth]{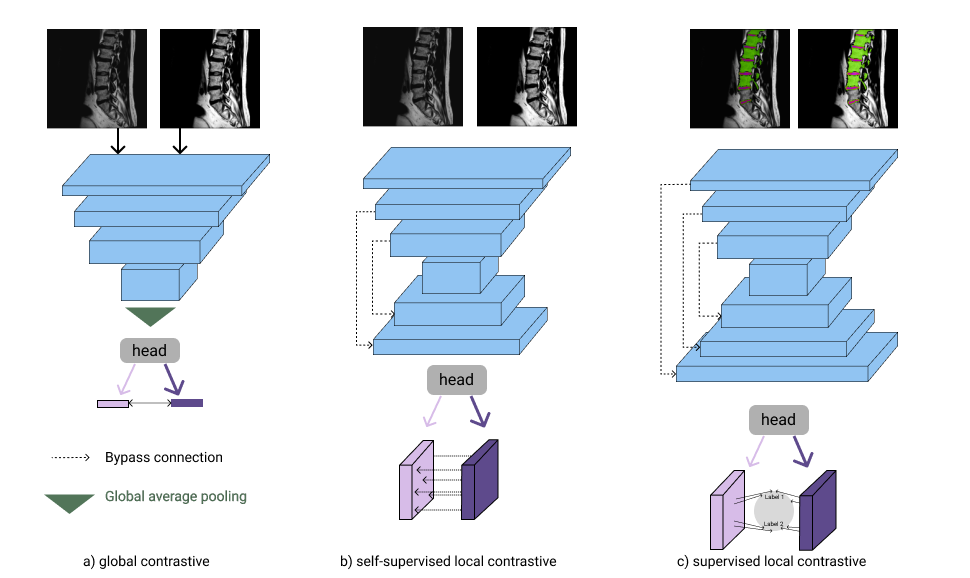}
\caption{Different examples of constructive learning architectures~\citep{hu2021semi}.}
\label{R3_fig20}
\end{figure*}

In addition, local contrastive learning can help students learn distinctive representations of local regions instead of relying on a global representation. Suitable for image segmentation. Figure~\ref{R3_fig20} shows contrastive learning architectures.

\subsection{Validation and Evaluation of SSMIS}

Semi-supervised learning for medical image segmentation is typically validated using standard protocols similar to fully supervised approaches, with careful comparisons to fully-supervised baselines. Researchers often partition the limited labeled data into training and validation sets (or use cross-validation) to tune models, and evaluate final performance on a held-out test set or challenge benchmark. A common strategy is to vary the fraction of labeled data and compare semi-supervised models against fully supervised models trained on the same subset of labels. This reveals how close the semi-supervised method can approach the upper bound of using all labels. Semi-supervised models are expected to match or exceed such metrics relative to low-data baselines. Benchmark datasets are frequently used for evaluation enabling direct comparison to fully supervised competitors. Authors often report that semi-supervised models narrow the gap to fully-supervised performance as unlabeled data are leveraged, sometimes even outperforming a fully-supervised model trained on the same small labeled set~\citep{hooper2023evaluating}. Ultimately, rigorous validation on independent data (e.g. from external institutions or challenge leaderboards) is crucial to demonstrate that semi-supervised segmentation methods maintain accuracy and generalize in clinical scenarios.

While existing SSMIS techniques have achieved comparable results with fully supervised frameworks in specific contexts, they still suffer from certain limitations, such as misaligned distributions and class imbalance, uniform weight for unlabeled data, and integration with annotation-efficient approaches. In SSMIS, the size of unlabeled data far exceeds that of labeled data, and they may follow different statistical distributions, which could diminish performance \citep{jiao2023learning}. Furthermore, the trained model can exhibit bias toward majority classes and, in some cases, completely ignore minority classes \citep{hou2022semi}.


The traditional approach to SSMIS involves supervised loss for labeled data and unsupervised loss/constraints for unlabeled data. In most cases, there is a single weight to balance between supervised and unsupervised loss, which overlooks that not all unlabeled data is equally appropriate for the learning procedure \citep{jiao2023learning}. This could be overcome by assigning individual weights to each unlabeled example to encourage the model to exploit more helpful information from unlabeled data \citep{ren2020not}.

\begin{figure}
\centering
\includegraphics[width=0.48\textwidth]{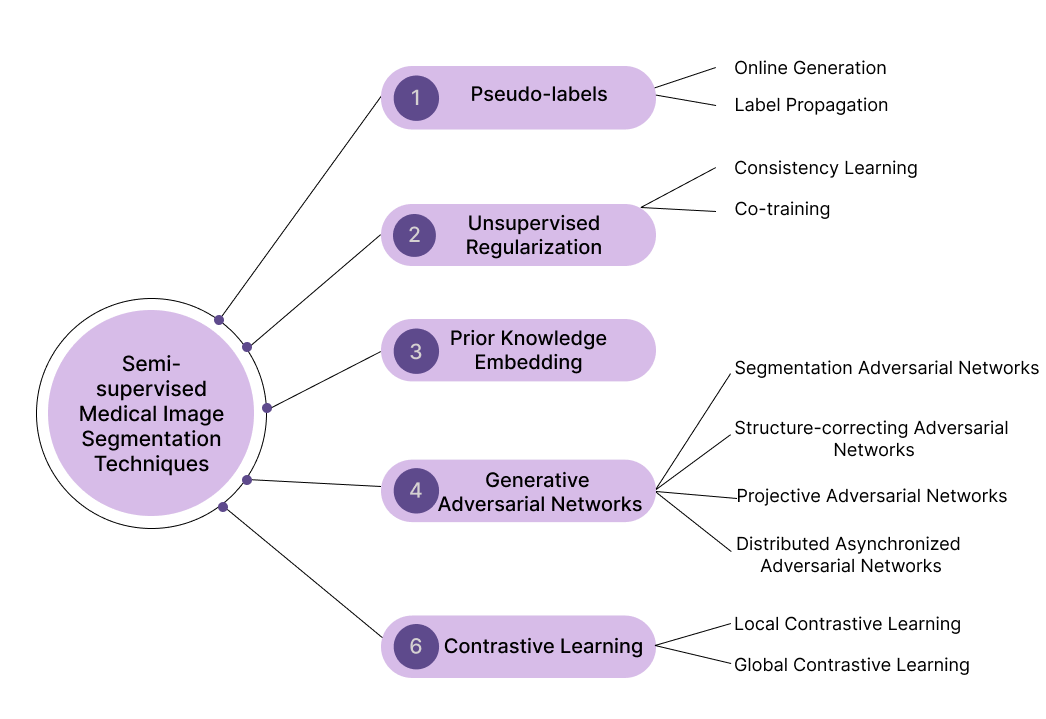}
\caption{Summary of Semi-Supervised Medical Image Segmentation Techniques.}
\label{R3_fig21}
\end{figure}

Since acquiring fully annotated medical images is a complex process, some contributions suggest integrating SSMIS with other annotation-efficient approaches that utilize partially labeled datasets \citep{zhang2021automatic}, leverage box-level or pixel-level annotations \citep{zhang2022cyclemix} or exploit noisy labeled data \citep{xu2021noisy}. Furthermore, it is suggested that SSMIS be integrated with few-shot segmentation to improve the generalization ability to segment unseen images better. The recent introduction of SAM (segment anything model) \citep{kirillov2023segment}, which serves as a pseudo-label generator for image segmentation, promises future development in the field of SSMIS. Figure~\ref{R3_fig21} summarizes different SSMIS techniques

\section{Recent and Emerging Trends in Medical Image Segmentation}
\label{s6}

This section outlines recent innovations and future research directions in MIS. 

\subsection{Cascaded Networks}

Increasing the network depth to utilize larger receptive fields is unsuitable for memory and computational reasons, particularly in the context of medical images' volumetric (3D) nature, and with the added downside of high-resolution detail loss as network depth increases. To overcome this problem, the idea is to utilize cascaded networks in a pyramid-like structure that performs high-resolution segmentation while also considering contextual information from lower resolutions \citep{conze2023current}. The weights of the lower resolution model are used as initializers of the higher resolution model through transfer learning (to be detailed subsequently in this section). There are three types of cascaded networks: Coarse-fine segmentation, Detection segmentation.


In coarse-fine segmentation, two 2D convolution networks are cascaded together; the first performs coarse (low-resolution) segmentation, and the second uses the output of the first to achieve fine segmentation \citep{wang2022medical}.
The first network model (possibly RCNN \citep{he2017mask} or You-Only-Look-Once (YOLO) \citep{bochkovskiy2020yolov4} is used for target location identification. Another network is used for further detailed segmentation based on the results of the first network. 
Since 2D networks cannot learn temporal data in the third dimension and 3D networks are too expensive in computation and memory, pseudo-3D segmentation methods have been proposed that involve the cascading of multiple 2D networks together for a more efficient version of 3D networks \citep{wang2022medical}.  

\subsection{Attention Mechanisms}

Similar to the human visual system's tendency to focus on a tiny portion of highly relevant perceptible information, with disregard for other perceivable stimuli deemed irrelevant, DL frameworks use attention mechanisms to selectively focus on the more essential aspects of an image \citep{conze2023current}. The model adaptively weighs its obtained features to focus specifically on those needed for the analytical task, suppressing feature responses in irrelevant backgrounds. The attention problem can be formulated using query (the target image), key (plausible target features), and value (the best matching regions). 

\begin{figure*}
\centering
\includegraphics[width=0.9\textwidth]{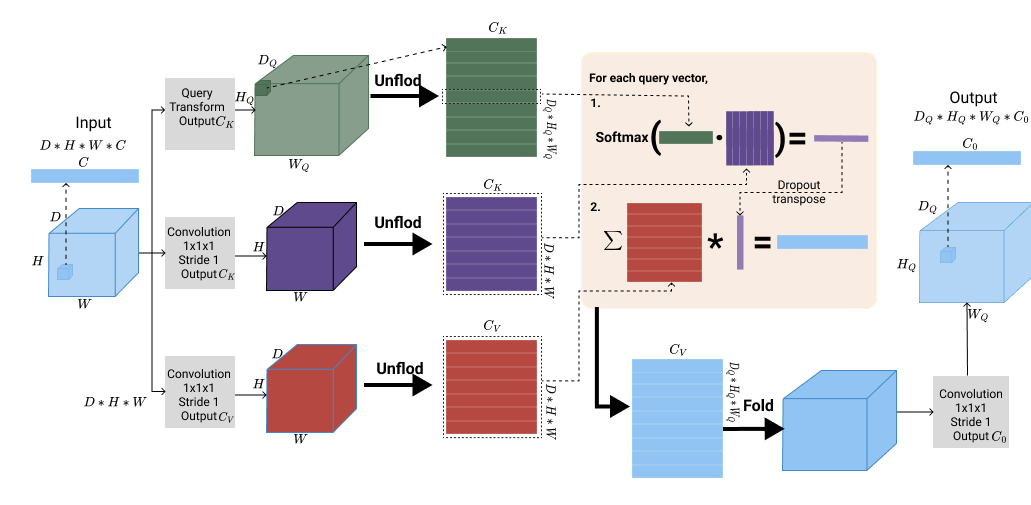}
\caption{Global Aggregation block in non-local U-net \citep{wang2020non}.}
\label{R3_fig22}
\end{figure*}

The channel attention framework is formulated as each channel represents a feature map that typically denotes distinct objects. Consequently, channel attention strives to calibrate the weight of each channel, selecting the entities that deserve more attention. The squeeze and excitation block \citep{iantsen2021squeeze} is an excellent example of Channel Attention, where global spatial information is captured through a squeeze operation (like global average pooling), and an alignment function (excitation module) captures channel relationships. It outputs an attention vector using fully connected and non-linear layers followed by a sigmoid function. Finally, each channel of the input feature map is multiplied by the corresponding element in the attention vector for contextualization. 
Similar to Channel Attention, Spatial Attention attempts to adaptively calibrate the weight of each part of the image, choosing where to focus its attention using an adaptive area selection procedure \citep{conze2023current}. 
Mutual attention between different sequences of MRI leads to improvement of low-grade brain tumors~(\citep{s24237576}). The branch Attention framework separates the attention problem into multiple sub-modules (branches), where each branch focuses on a particular aspect and exchanges significant information with other branches. 
Spatial Attention ignores inter-channel information variation, and Channel Attention pools global information directly; therefore, mixed attention networks have been designed to combine both advantages in \citep{wang2022medical}.
In \citep{wang2020non}, a non-local U-net was implemented with a self-attention mechanism and a global aggregation block that extracts full image information during up-sampling and down-sampling. The non-local block is shown in Fig.~\ref{R3_fig22}. 

\subsection{Transformer-based Methods}

\begin{figure*}
\centering 
\includegraphics[width=0.9\textwidth]{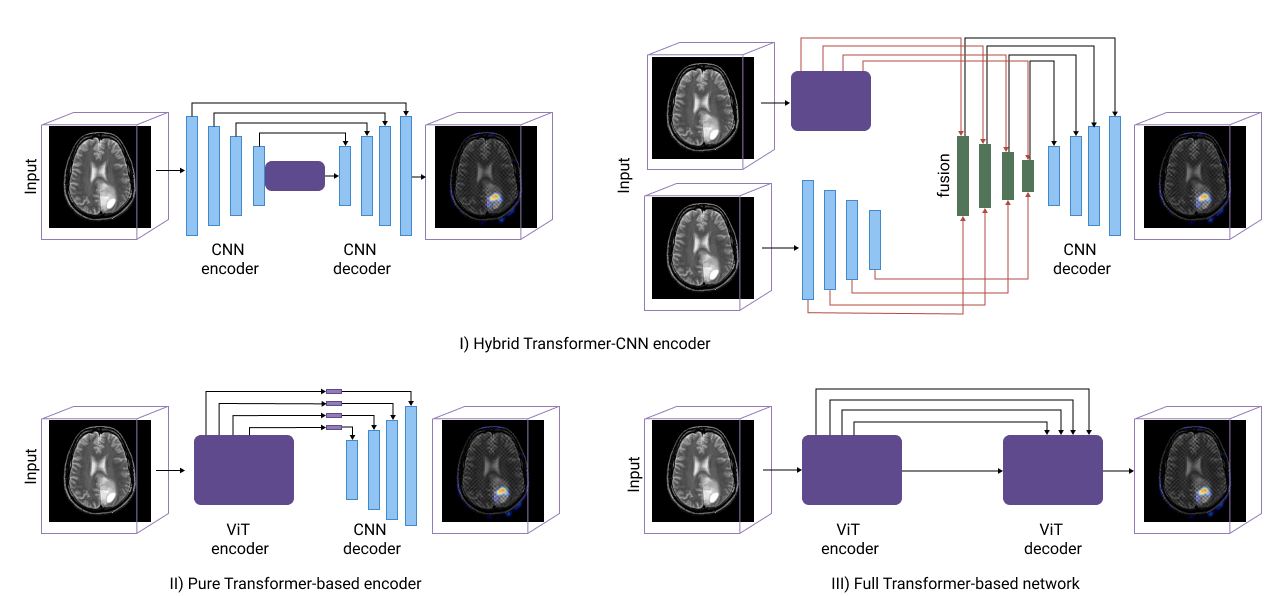} 
\caption{Transformer architectures. I) Hybrid Transformer-CNN encoder, II) Pure Transformer-based encoder, and III) Full transformer-based network \citep{conze2023current}.} 
\label{R3_fig23} 
\end{figure*}

Transformers are attention-based architectures that were originally developed for natural language processing. Over time, they have been adapted for use in image processing tasks as well. A major milestone in this transition was the introduction of Vision Transformers by~\cite{dosovitskiy2020image}. In this work, the authors explored replacing convolutional neural networks with purely Transformer-based, convolution-free models for image segmentation. Unlike CNNs, Vision Transformers (ViT) offer a complete view of a single layer and facilitate parallel processing. The transformer-based encoder is a sequence of alternating layers of multi-head self-attention (MSA) and multilayer perceptron blocks (MLP). A layer normalization is applied before each block, and a residual connection is applied after \citep{conze2023current}. In MIS, Transformer-based segmentation models still adopt a U-net-shaped architecture, as shown in Fig.~\ref{R3_fig23}.

A pure Transformer-based encoder enable the global context modeling capabilities of Transformers to effectively capture relationships between spatially distant voxels in medical images. When combined with convolutional upsampling layers and multi-level feature aggregation, this architecture significantly improves segmentation performance \citep{Yang2024, Sui2025}. Various implementations have emerged to optimize these models for dense prediction tasks, such as segmentation. For instance, hierarchical Vision Transformers (ViTs) are designed to extract features at multiple resolutions without being limited to fixed subregions of an image \citep{hatamizadeh2021swin, dong2021polyp}. 
One notable example is the 3D U-shaped model, which utilizes nested hierarchical Transformers and incorporates global self-attention within non-overlapping blocks \citep{yu2022characterizing}. This design streamlines the model while still effectively capturing both local and global contextual information, making it a powerful tool for medical image analysis.

\begin{figure*}
\centering 
\includegraphics[width=0.9\textwidth]{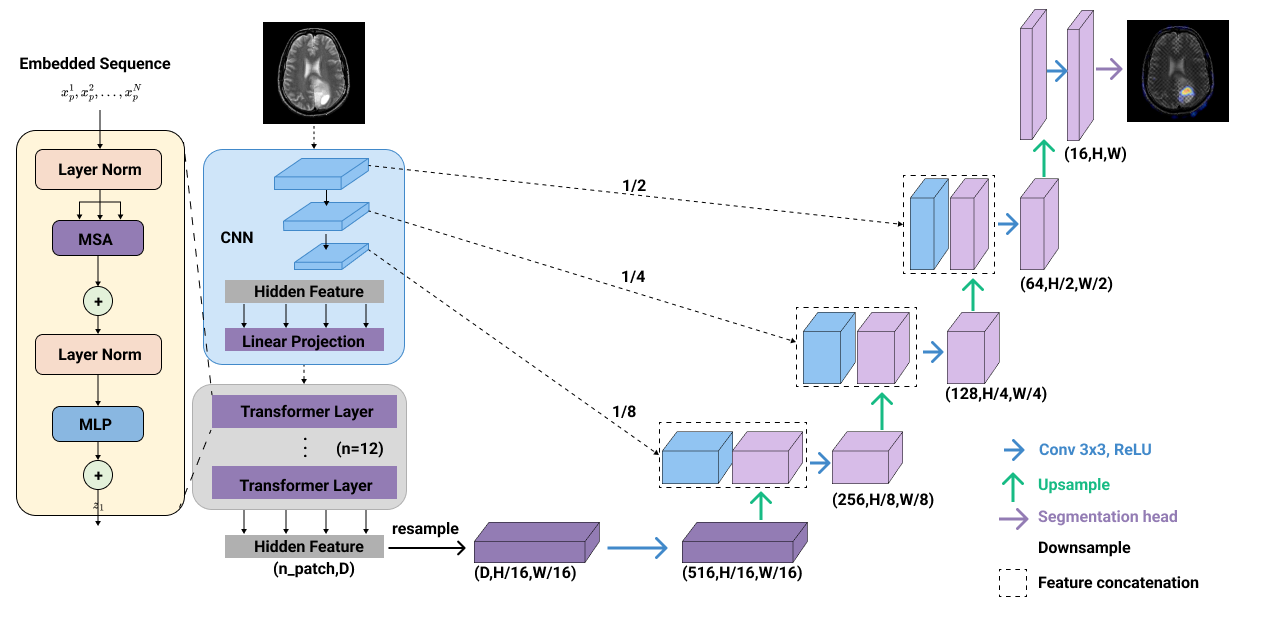} 
\caption{TransUnet architecture \citep{chen2021transunet}.} 
\label{R3_fig24} 
\end{figure*}

This framework represents Transformers' global context modeling capabilities but with a CNN inductive bias, in that CNNs stack convolution blocks that capture multi-scale context feature maps. In contrast, Transformers capture long-term dependencies among features that can be lost in purely Transformer-based models. Finally, the Transformer output is up-sampled into a 4D feature map in a CNN-based decoder to recover the full segmentation mask \citep{conze2023current}. This approach is illustrated in Fig.~\ref{R3_fig24}.

Transformer-based architectures have shown promise in medical image segmentation, but scaling them to large 3D images and clinical datasets poses technical challenges. Memory consumption is a primary concern: the self-attention mechanism has quadratic complexity in the number of input tokens (patches). Medical images (especially 3D volumes) produce far more patches than typical 2D natural images, making a vanilla ViT infeasible without downsampling. Early attempts to apply ViT directly in segmentation resulted in poor accuracy unless heavy downsampling was used. For example, the TransUNet framework addresses this by using a CNN encoder to first reduce spatial resolution, then feeding tokens to a ViT~\citep{chen2021transunet}. Similarly, Swin-UNet adopts windowed attention (local self-attention) to curtail memory usage, computing self-attention within small patches instead of globally~\citep{cao2022swin}. Such hierarchical Transformers with local attention greatly reduce memory and FLOPs while approaching the global context modeling of full attention. Even with these optimizations, training speed and model complexity remain concerns. Transformer models often have tens of millions of parameters and require more computation than CNNs for the same input size. In one comparison, pure Transformer encoders (ViT) had significantly higher FLOPs than U-Net or hybrid models. Frameworks like UNETR~\citep{Hatamizadeh2022WACV}, which uses a ViT as the sole encoder, can exceed 100 million parameters, necessitating multiple high-end GPUs or large memory for training.

The full Transformers network is where Transformers are stacked end-to-end. Neural Architecture Search (NAS) \citep{weng2019unet} strives to automate the iterative process of network design, traditionally executed manually by researchers \citep{conze2023current}. It is a field that overlaps with hyperparameter optimization \citep{ha2019bayesian} and meta-learning \citep{vanschoren2018meta}. The application of NAS to the design and optimization of Transformer-based architectures for medical image segmentation. Given the increasing complexity and scale of Transformer models (e.g., ViT, TransUNet, UNETR), manually designing optimal architectures becomes both time-consuming and prone to suboptimal performance or overfitting. NAS offers a systematic and automated way to explore and
optimize design spaces. There are three NAS domains in contemporary research: Search space, search strategy, and performance estimation. This comprises the collection of existing networks to be searched. A global search space represents the search for an entire network structure, whereas a cell-based search space pursues only a few small structures stacked together to form more extensive networks \citep{wang2022medical}. This framework attempts to execute the fastest possible search within the search space. Possible search strategies include reinforcement-based learning, evolutionary algorithms, and gradients \citep{wang2022medical}.

These are fixed quantifications of performance and rarely vary among different NAS algorithms. The nnUnet \citep{isensee2021nnu} is an example of a NAS-based product. Authors argue that excessive manual adjustment to network structure may lead to overfitting and, therefore, offer a new framework that adapts itself to any new dataset. It focuses on preprocessing (resampling and normalization), training (loss, optimization settings, data augmentation), inference (test-time-augmentations and model integration), and post-processing (enhanced single-pass domain) \citep{wang2022medical}.

\subsection{Cross-modality segmentation and fusion techniques}

\begin{figure*}
\centering
\includegraphics[width=0.9\textwidth]{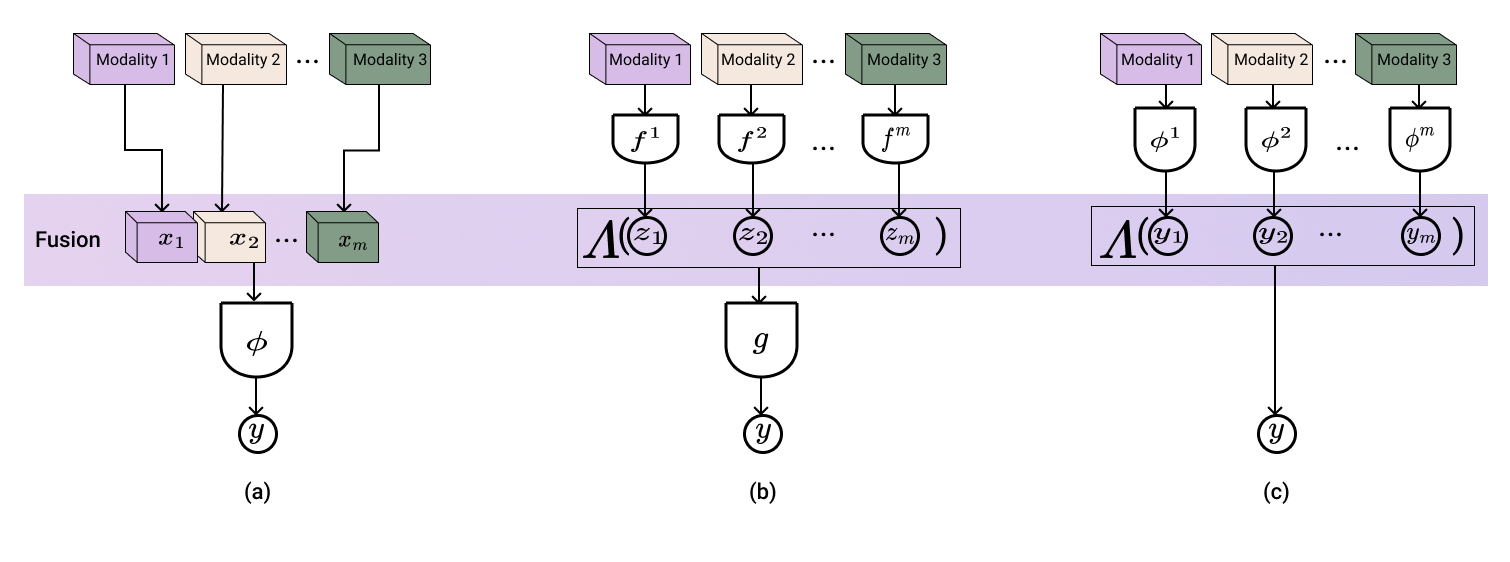}
\caption{Cross-modal framework for paired data. a) Early fusion, b) mid-fusion, c) late fusion.}
\label{R3_fig25}
\end{figure*}

Cross-modal information is a potentially valuable but largely unused feature in MIS. However, exploring complementary and redundant information across various imaging modalities can improve segmentation performance. Data could be paired (coming from the same patient) or unpaired, which determines the type of fusion strategy to be adopted (early, mid, or late fusion). This can be observed in Fig.~\ref{R3_fig25}. 

The most straightforward strategy is the early fusion strategy, where the modalities are integrated at the input level; this has the advantage of simplicity, which allows for more complex segmentation strategies like GAN-based approaches. However, early fusion strategies cannot understand non-linear relationships between low-level features from different modalities, especially when they have significantly different statistical properties.
In paired data and mid-fusion strategy, multi-modal data is separately processed in different paths, then mapped into a common latent space via a fusion operation, and finally input into a decoder; the target is to emphasize the most significant features across modalities \citep{conze2023current}.
 \begin{itemize}
     \item[(a)] Single layer fusion, where each modality has its own encoder, without inter-encoder communication, and a shared decoder. Encoded data is fused by concatenation, addition, or convolution. This technique encourages the network to learn the most correlated features across modalities and the most useful spatial information. Still, the single level of abstraction prevents it from learning within and between modalities.
     \item[(b)] Multi-layer fusion extends the idea of residual learning by utilizing skip connections that bypass spatial features between modalities. Thus, lower and higher-level features are fused at different abstraction levels, increasing the network's ability to capture complex cues across modalities. Fusing multi-modal contextual information at multi-layer stages is further facilitated by employing attention mechanisms that bridge early feature extraction and late decision-making. 
 \end{itemize}

\begin{figure*}
\centering
\includegraphics[width=0.9\textwidth]{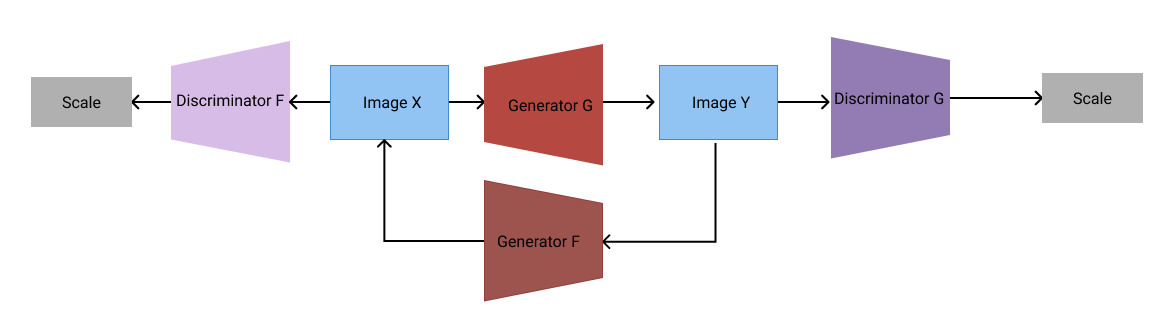}
\caption{The Cycle GAN architecture \citep{zhu2017unpaired}.}
\label{R3_fig26}
\end{figure*}

When working with paired data in medical image segmentation, late fusion is a common approach. In this strategy, individual segmentation branches, each processing a different data modality, are integrated during the decoding stage. This involves mapping all computed feature maps into a single, unified feature space using operations like concatenation, averaging, or weighted voting, followed by subsequent convolutional layers to refine the combined information \citep{zhang2021deep}. Both mid- and late-fusion strategies often lead to improved performance because each modality is fed into a separate network, enabling the model to learn complex and complementary feature information unique to that data type. This advantage, however, comes with a trade-off: using multiple networks significantly increases the memory and computational power required \citep{conze2023current}.
A significant challenge in medical imaging is the difficulty and high cost of collecting large datasets of paired images. This has led to a growing focus on utilizing unpaired datasets. When working with such data, domain adaptation becomes crucial. This specialized area within transfer learning addresses situations where the categories or labels are consistent across different data domains, but the characteristics of the domains themselves vary. The primary goal is to find a suitable transformation that allows a segmentation model, initially trained on a source domain with labeled data, to perform effectively on a target domain, which might contain unlabeled or differently acquired data.
Domain adaptation is categorized by the availability of labeled data in the target domain: it can be supervised (both labeled source and target data are available), semi-supervised (labeled source data is available, with partial labeled target data), or unsupervised (only labeled source data is available, with no labeled target data). Cycle Generative Adversarial Networks (CycleGANs) are frequently employed in unsupervised domain adaptation \citep{wang2022medical}. As illustrated in Fig.~\ref{R3_fig26}, a CycleGAN typically uses two generators: one translates an image from a source domain to a target domain, and the other translates the output back, ensuring that the learned transformations preserve the essential content of the images while adapting their style or characteristics to the target domain.

\begin{figure*}
\centering
\includegraphics[width=0.9\textwidth]{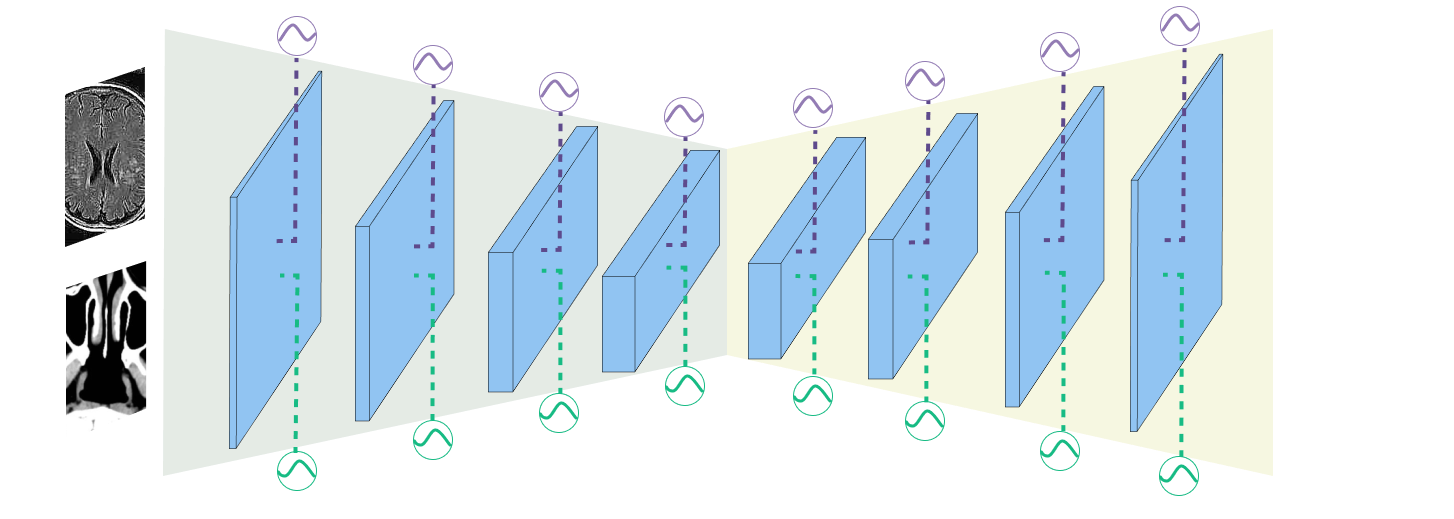}
\caption{Shared convolutional kernels and domain specific feature normalization \citep{dou2020unpaired}.}
\label{R3_fig27}
\end{figure*}

The inter-modality variance problem has been addressed; we will also discuss the multi-domain segmentation as an intra-modality variance problem. For example, two different CT scanners can vary intensity distributions in their output scans. This is particularly relevant if the training data set prepares the model for generalized evaluations on different data sets (different instances of the same imaging modality). The underlying assumption is that extraction of robust domain-invariant feature representations is possible if the redundancy between multiple-intensity domains is adequately exploited, enabling the model to perform better than a domain-specific model. Differences in imaging systems, reconstruction settings, and acquisition protocols make multi-domain segmentation essential in real-life medical applications \citep{conze2023current}. Attempts to achieve multi-domain segmentation include using adversarial networks to learn domain-invariant features~\citep{kamnitsas2017unsupervised} and using transfer learning~\citep{karani2018lifelong} with a single encoder-decoder segmentation network with shared convolutional kernels but domain-specific feature normalization (Fig.~\ref{R3_fig27}).

\subsection{Distributed Learning Frameworks}
 
Distributed Learning refers to techniques and paradigms that enable deep learning models to be trained across multiple computing nodes or data sources, rather than on a single, centralized dataset. These approaches are critical for unlocking the full potential of deep learning in clinical settings, allowing for robust model development even when data cannot be centrally pooled. Distributed learning in medical image analysis offers several key methodologies, including federated learning for privacy-preserving collaborative model training, transfer learning and domain adaptation for using existing knowledge across different datasets, and knowledge distillation and multi-task learning for efficient model deployment and comprehensive learning from diverse tasks.

Transfer learning is beneficial in a limited datasets context; this strategy entails pre-training the model using existing large-scale datasets before applying it to the problem at hand~\citep{sistaninejhad2023review}. Pretrained models include LeNet-5~\citep{lecun1998gradient}, AlexNet~\citep{krizhevsky2017imagenet}, Visual Geometry Group (VGG Net) \citep{simonyan2014very}, GoogleNet \citep{szegedy2015going}, ResNet~\citep{he2016deep}. These models are then fine-tuned to MIS by freezing the convolutional base and training some layers only \citep{sistaninejhad2023review}. However, domain adaptation is still required when moving from natural to medical images; furthermore, pretrained models often rely on 2D datasets, and their applicability to 3D data is limited~\citep{wang2022medical}.

Training different models on the same data and then averaging their predictions is a feasible but computationally expensive strategy; compressing their combined knowledge into a single model through knowledge distillation is much easier to implement~\citep{hinton2015distilling}. The idea is to transfer information from a well-trained teacher network to a lightweight and compact student network to improve the performance of the student model. Extensions of this framework include the multiple teacher's single student model (MTSS)~\citep{zhang2022unsupervised}, which attempted to preserve patient privacy by constructing a multi-organ segmentation student network that learns from multiple pre-trained single-organ segmentation models. 


Multi-task learning strives to utilize information shared across two or more auxiliary tasks to improve the handling of each task~\citep{tajbakhsh2020embracing}; viewed as an inductive transfer process, the introduction of an inductive transfer bias allows the prioritization of specific hypotheses over others~\citep{ruder2017overview} toward more generalizable solutions. It can leverage a variety of heterogeneous forms of annotations to solve several image-processing tasks at once, using a cascade of task-specific sub-networks or networks with shared encoders and task-specific decoders~\citep{playout2018multitask} to benefit from partial parameter sharing between tasks. 

\begin{figure*}
\centering
\includegraphics[width=0.9\textwidth]{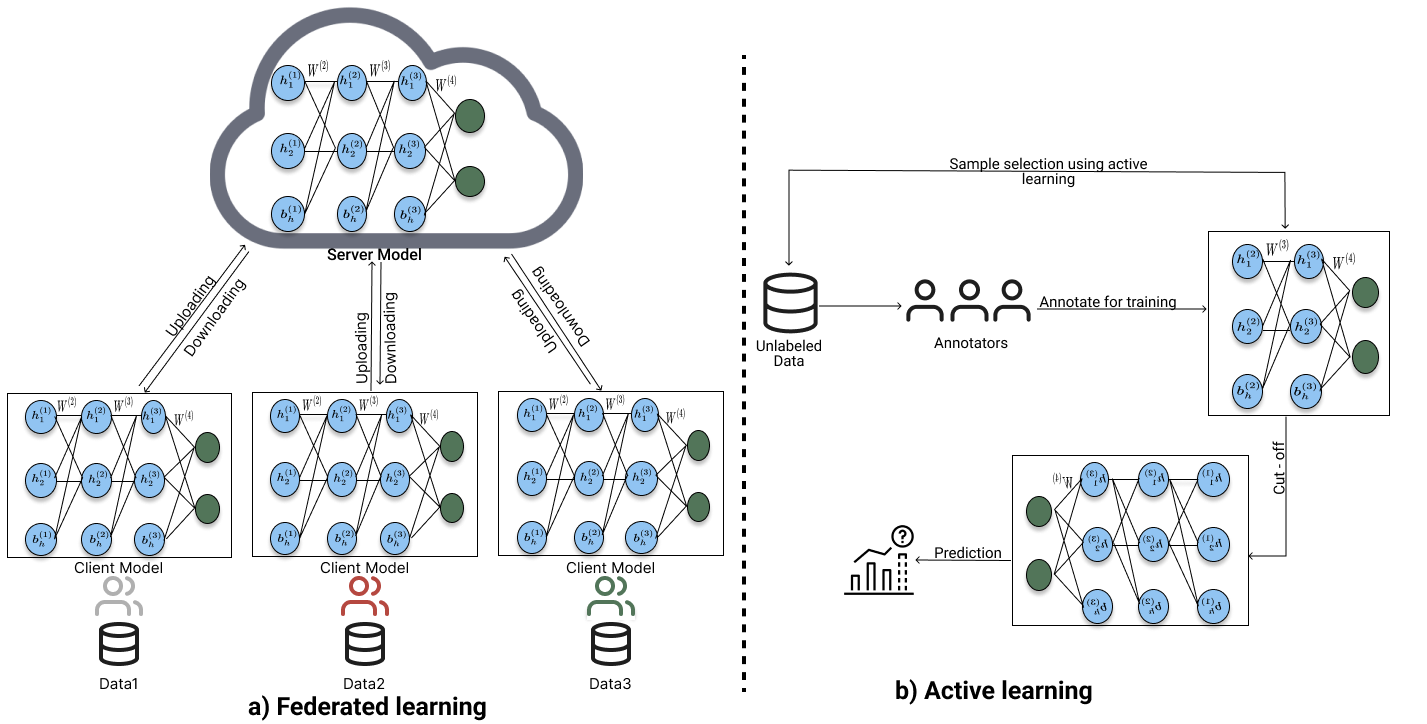}
\caption{General framework for (a) federated learning and (b) Active Learning \citep{conze2023current}.}
\label{R3_fig28}
\end{figure*}

Federated learning enables the distributed training of deep models without sharing data between multiple institutions to protect patient privacy. Each institution maintains an individual model that focuses on the local data and requests a global model from a central server to download the global model weights. During training, local model weights are sent to the central server for updating. The central server aggregates the feedback from individual institutions, and the global weights are then updated according to predefined rules based on the varying feedback quality from individual institutions. The model is shown in Fig.~\ref{R3_fig28} (a). This is a promising research direction in that it allows for the generation of larger training data sets while maintaining patient privacy~\citep{conze2023current}.

\subsection{Active Learning}

Active learning frameworks offer a promising solution to the challenges of annotating large medical datasets by iteratively selecting the most valuable samples for training deep learning models. This approach not only reduces the annotation burden but also enhances model performance, making it a valuable tool in the development of efficient and accurate medical AI systems \citep{Nath20212534,Khanal202448,Carse201920}.
Active learning also allows the model to focus on the most challenging samples, thus improving its performance. However, this choice process is challenging and involves uncertainty and representative criteria \citep{budd2021survey}. The framework is presented in Fig. \ref{R3_fig28} (b).


\subsection{Segmentation Uncertainty}
Various sources of uncertainty impacting network performance can be grouped into epistemic and aleatoric uncertainty.
Epistemic (Model) Uncertainty describes parameter uncertainties in the model due to insufficient training. It is reduced by providing more training time and data. Monte-Carlo dropout (or Test Time Dropout) is a stochastic technique that yields an epistemic uncertainty map dependent on dissimilarity in predictions. However, this dropout may negatively affect the training performance itself. Other epistemic uncertainty quantification techniques include deep ensembling, where independent networks are trained, and their predictions are averaged together to obtain uncertainty maps \citep{conze2023current}.
 
Aleatoric Uncertainty relates to the inherent uncertainty of the data itself, which can be further subcategorized into homoscedastic uncertainty (constant for all samples in a given set) and heteroscedastic uncertainty (varying among samples). Homoscedastic uncertainty stems from physical properties of the imaging modality, like the variation in positron range and Compton scattering. In contrast, heteroscedastic uncertainty may be due to heterogeneity in annotation quality \citep{castro2020causality}. Aleatoric uncertainty can be quantified through Test-Time Data Augmentations (TTA) in which multiple forward passes are performed to augmented inputs \citep{Conde2022}. This approach enhances model performance and provides a robust measure of uncertainty, making it valuable for various MIS tasks.

\begin{figure*}
\centering
\includegraphics[width=0.9\textwidth]{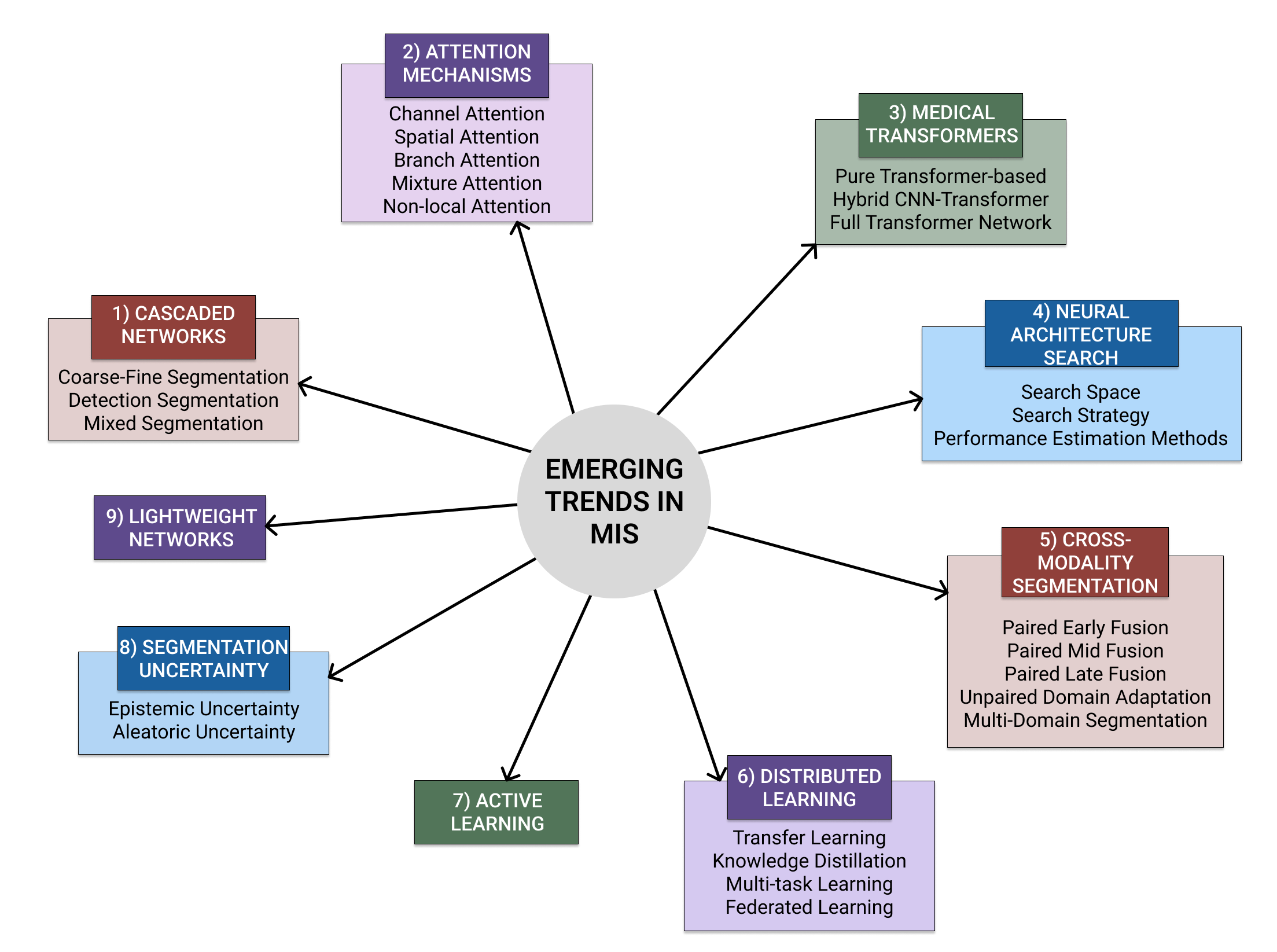}
\caption{Summary of Emerging Trends.}
\label{R3_fig29}
\end{figure*}

Applying CNNs or Transformers for 3D MIS is a computationally expensive process involving many parameters to train \citep{ma2022fast}. The computational resources available on specific devices may be inadequate for executing complex 3D segmentation tasks. Consequently, there has been a growing interest within the research community in developing lightweight deep learning models capable of performing efficient 3D segmentation under limited computational conditions. Some solutions include depth-separable convolutions \citep{lei2021automatic}, which involve replacing a 3D convolution kernel 3$\times$3$\times$3 with 1$\times$3$\times$3 intra-slice and 3$\times$1$\times$1 interslice convolutions, combinations between pointwise, group-wise, and dilated convolutions \citep{chen20193d}, or channel reduction \citep{yaniv2020v}. It is worth nothing that model size is particularly relevant in uncertinity estimation because they are often deployed in resource-constrained environments, where uncertainty-aware prediction are crucial for safe decision-making.
While heavy-weight models generally offer higher performance due to their extensive parameter counts and computational resources, lightweight models have significantly improved competitive accuracy and efficiency. The trade-off between performance and computational complexity can be effectively managed through innovative architectural designs, attention mechanisms, and efficient feature-processing techniques. Lightweight models like MediLite3DNet, SegFormer3D, and HL-UNet demonstrate that it is possible to achieve high performance in 3D medical image reconstruction without the heavy computational burden typically associated with larger models\citep{Qin2024451, Dai2024, Liu2025}. Figure~\ref{R3_fig29} presents a summary of emerging trends.

\section{Case Study: Lumbar Spine Segmentation}
\label{s7}

Lumbar spine segmentation has seen significant advancements through various methodologies, particularly deep learning and hybrid approaches, which have demonstrated high accuracy and robustness across different imaging modalities. These advancements are crucial for enhancing clinical diagnostics and treatment planning. Lumbar spine segmentation applications include:
detection of spinal anomalies, surgical planning, and treatment monitoring \citep{lu2023lumbar, Basak2025, He2024}. Another application is diagnosis of lumbar deformities and fractures \citep{Mushtaq2022}.
Lumbar spine segmentation has common challenges as 
overlapping shadows in X-ray images, unclear boundaries, and inter-patient variability \citep{Kim2021}.
High similarity between vertebral bones and intervertebral discs in MRI images \citep{Liu202477999}.

\subsection{Overview}

The vertebral column is essential in supporting the human body and protecting the spinal cord and, subsequently, the central nervous system. The spinal region, located therein, comprises the cervical, thoracic, and lumbar spine \citep{andrew2020spine} and plays a primary role in mobility in the musculoskeletal system, in addition to sustaining organ structure and protecting the body from shock \citep{qadri2023ct}. Defects within this region result in Vertebral Misalignment, the disease responsible for chronic back pain, which is construed as one of the most prevalent medical problems in contemporary society \citep{andrew2020spine}. The slightest damage to this region risks causing immense pain and bodily malfunction to the afflicted patient due to the large concentration of nerves therein \citep{andrew2020spine}. Different types of abnormalities in this region \citep{raghavendra2018automated}, some of which require emergency treatment, like osteoporotic fractures.
In contrast, others, like disc degeneration and scoliosis, can be therapeutically treated, though equally painful in their effects on the human body. Biomechanical changes can result in disability
and severe discomfort in the short term. Still, they can have far worse long-term complications, such as an eightfold higher mortality rate due to osteoporosis \citep{qadri2023ct}. Yet, it remains underdiagnosed despite its critical nature. To emphasize the impending criticality of the medical problem at hand, we summarize in Table \ref{tab01} the various diseases that manifest in the human body of a patient suffering from spinal malalignments \citep{andrew2020spine}. 

\begin{table*}
\caption{Manifestations of spinal malalignments.}
\begin{tabular}{|p{2.0cm}|p{3.5cm}|p{5.0cm}|p{4.0cm}|}
\hline
Disease	& Description & Cause & References \\
\hline
Lumbar Spine Stenosis & Narrowing of spine canal & Collisions of tissues located between vertebrae causing inflammation and pain in spinal nerves & \citep{webb2024lumbar, ross1968aortic, ghosh2012new}\\
\hline
Scoliosis & Spine assumes an S shape or C shape & Exaggerated neurological activity, birth defects& \citep{kumar2024scoliosis, aebi2005adult, horng2019cobb} \\
\hline
Osteoporotic Fractures & Degradation of bone density & Bones weaken over time, most commonly in elderly people aged 80+ & \citep{khan2024osteoporotic, johnell2005epidemiology}\\
\hline
Thoracolumbar Fractures & Injuries in thoracic and lumbar vertebrae & Hard fall or accident that induces intense impact on the back & \citep{azizi2024thoracolumbar, raghavendra2018automated, mcafee1983value}\\
\hline
Degeneration & Weakening of bones & Aging & \citep{fine2023intervertebral, lim2012age, jebri2015detection}\\
\hline
\end{tabular}
\label{tab01}
\end{table*}

In an attempt to diagnose the patient's spine, traditional methods involved inspection tools like magnetic resonance imaging (MRIs), X-rays, computed tomography (CT-scans), and positron imaging tomography (PET) scans \citep{jeon2020application, sneath2022objective}. However, these modalities heavily depend on radiologists with years of experience and are subject to diagnostic inaccuracies \citep{boulay2014pelvic}. The challenges to manual segmentation and analysis of medical images of the spine include the time-consuming nature of manual annotation \citep{andrew2020spine} as well as the inaccuracies exacerbated by the small-sized nature of the region of interest and the consequent difficulty faced in bounding it \citep{forsberg2017detection}. Furthermore, the inconsistent standards of different radiologists lead to significant differences in segmentation results \citep{lu2023lumbar}, and the traditional methods of region-based and threshold-based segmentation are limited by the varying influences of imaging equipment and principles, and the consequent complexity of the shape and content of medical images \citep{furqan2018automatic}. Thus, an imminent need for automated segmentation arises \citep{qadri2023ct, pereira2016brain} to facilitate the prompt morphological analysis and effective clinical treatment of these pathologies \citep{lu2023lumbar}. 

\subsection{Vertebral Segmentation Literature Review}

Automatic vertebral segmentation can be perceived as a pixel-level classification method \citep{lu2023lumbar} with diagnostic significance in estimating spinal curvature and recognizing spinal deformities \citep{qadri2018deep} as well as facilitating finite element modeling analysis, biomechanical modeling, and surgical planning for metal implantations. Traditionally, automatic segmentation was achieved using prior-shape models like statistical shape models \citep{pereanez2015accurate, castro2015statistical, rasoulian2013lumbar}, geometric models \citep{ibragimov2017segmentation, ibragimov2014shape}, Markov Random Fields (MRF) \citep{kadoury2013spine, kadoury2011automatic} and active contours \citep{athertya2016automatic}; all of which essentially revolved around fitting a shape before the spine and distorting it into conformity to the spinal shape \citep{qadri2023ct}. Other models include a priori variational intensity models \citep{hammernik2015vertebrae}, level sets \citep{lim2014robust} as well as landmark-framework-based automatic segmentation models \citep{korez2015framework}.

The proliferation of machine learning techniques led to numerous contributions in lumbar spine image segmentation. In \citep{suzani2015deep}, vertebral structures were identified using a multi-layer perceptron (MLP) and segmented by deformable registration. In \citep{chu2015fully}, vertebrae were located using random forest regression and then subsequently segmented at the voxel level utilizing a random forest classifier. Authors of \citep{sneath2022objective} proposed a two-stage decision forest coupled with a morphological image processing technique that facilitates the automatic detection and identification of vertebral bodies in arbitrary field-of-view volume CT scans. In \citep{korez2016model}, the deformation model was combined with convolutional neural networks (CNNs) to learn spine features and output the probability map of the spine through the CNN, thereby guiding the deformation model to create a boundary for the spine and, therefore, realize the objective of spine segmentation. In \citep{lang2019differentiation}, the authors combined SVM with the histogram of oriented gradients (HOG) methodology to create tight bounding boxes that encapsulate the region of interest in the upper vertebrae in a mid-sagittal MRI image. Discs are initially segmented based on corresponding axial MRIs calculated from the intersection of the axial slices with the Sagittal, and then leftover discs are segmented using a two-stage classifier. 
However, in recent years, the explicit modeling of vertebral shape gave way to data-driven learning techniques facilitated by the emergence of more sophisticated spine image datasets that have become publicly available \citep{lu2023lumbar}. The most prominent deep-learning-based image segmentation techniques depend on convolutional neural networks (CNNs) or specifically stacked sparse auto-encoders (SSAEs). 

CNN-based segmentation architectures include fully convolutional neural networks (FCN) \citep{long2015fully}, SegNet \citep{badrinarayanan2017segnet}, UNet \citep{ronneberger2015u} and 3D Unet \citep{cciccek20163d}. In \citep{al2018segmentation}, patch-based segmentation is employed to differentiate between the anterior and posterior parts of the spine. The image input layer receives each pixel in the patch and applies data normalization. The output is then fed into convolutional layers with equal-sized kernels equipped with a trained classification feature. Segmentation training is sped up using batch normalization, the ReLU function, and fully connected layers, which extract the final features. In \citep{al2019boundary}, Segnet is employed, and an encoder-decoder architecture encapsulates a sequence of convolution layers. The encoder selects image features at varying resolutions; the final output is a boundary detail. The convolution layers are equipped with filter banks, and the output is fed into batch normalization followed by a ReLU activation function.
Furthermore, max-pooling is used to implement subsampling. The decoder is responsible for upsampling the output signal and restoring it using the max pool layer. The result is then fed into a filter-equipped convolutional layer, followed by batch normalization and ReLU activation again. The final output is applied to a softmax function, and the result is the segmented image. 
In \citep{sekuboyina2017attention}, a CNN was utilized for spine localization and another CNN for spine segmentation. The input to the localization CNN is a 2-D slice of the spine, and they down-sample the real spine mask to obtain the localization network's ground truth, using patches to segment the vertebrae one at a time. Authors in \citep{janssens2018fully} used two 3D FCNs for spine localization and segmentation; the spine is localized using a bounding box obtained by regression in the localization network, and voxel-level multiclassification is performed in the segmentation network. In \citep{lessmann2019iterative}, only one 3D FCN is used for spine segmentation; an iterative method is employed to segment the vertebrae sequentially according to the prior rules of their appearance. A memory component is added to the network to ascertain that vertebrae in the current block have been segmented. Finally, a classification component is utilized to label the vertebrae that have been segmented. However, this method struggles to solve the problem of overlapping vertebrae \citep{lu2023lumbar}. 

\begin{table*}
\centering
\caption{Summary of recent methods for vertebra segmentation and analysis.}
\resizebox{\textwidth}{!}{%
{\begin{tabular}{|p{4cm}|p{6cm}|p{3cm}|p{4cm}|p{5cm}|}
\hline
\textbf{Method} & \textbf{Description} & \textbf{Datasets} & \textbf{Applications} & \textbf{Performance Metrics} \\
\hline

Patch-based Deep Learning (SSAE) & Divides 2D CT slices into patches, uses SSAE for feature extraction, and RUS for data balancing. & VerSe, CSI-Seg, Lumbar CT & Clinical applications & Precision: 89.9\%, Recall: 90.2\%, Accuracy: 98.9\%, F-score: 90.4\%, IoU: 82.6\%, DC: 90.2\% \cite{2} \\
\hline
Coarse-to-Fine Method with HMC & Initial coarse shape followed by HMC segmentation without shape prior. & Standard and non-standard vertebrae datasets & Orthopedics, interventional procedures & Robust to shape and luminance changes \cite{3} \\
\hline
Semi-automatic (Boundary Classification and Mesh Inflation) & Uses boundary classification and mesh inflation for vertebra segmentation. & 11 lumbar datasets & Diagnosis of spine pathologies & Detection rate: 93\%, DSC: 78\% \cite{4} \\
\hline
Machine Learning (SVM) & Uses gradient orientation histograms and SVM for cervical vertebra detection in MR images. & 21 T2-weighted MR images & Diagnosis and therapy & Accurate despite severe noise \cite{5} \\
\hline
Deformable Model-based Segmentation & Focuses on thoracic and lumbar vertebrae using a deformable model. & 16 patient datasets & Virtual spine straightening & Applicable to scoliosis and bone metastases \cite{6} \\
\hline
Probabilistic Energy Functions & Models intensity, spatial interaction, and shape for segmentation. & Clinical CT images, phantom datasets & BMD measurements & Robust under various noise levels \cite{7} \\
\hline
Edge-mounted Willmore Energy and Probabilistic Model & Combines intensity and shape information for segmentation. & 40 clinical CT images, phantom datasets & BMD measurements & Higher accuracy than alternatives \cite{8} \\
\hline
Statistical Shape Decomposition and Conditional Models & Part-based statistical decomposition for detailed segmentation. & 30 healthy CT scans, 10 pathological scans & Image-guided interventions, musculoskeletal modeling & Point-to-surface error improvement: 20\% (healthy), 17\% (pathological) \cite{9} \\
\hline
Iterative Instance Segmentation (FCN) & Uses FCN with memory component for iterative vertebra segmentation and labeling. & Multiple datasets (CT and MR) & Spine analysis, abnormality detection & DSC: 94.9\%, Anatomical identification accuracy: 93\% \cite{lessmann2019iterative} \\
\hline
Statistical Shape Models (SSM) & Uses SSM for lumbar vertebra segmentation with B-spline relaxation. & 5 patient datasets & Not specified & Global mean DSI: 93.4\% \cite{Forsberg2015215} \\
\hline
Deep Learning (Res U-Net) & Combines Res U-Net with Otsu’s method for feature extraction. & VerSe’20 & Preoperative planning & DSC: 87.10\% \cite{qadri2023ct} \\
\hline
Interactive Segmentation (Grow-cut) & Semi-automated method using user-defined seed pixels. & 23 subject datasets & Diagnosis of herniated discs & Accuracy: 97.22\%, Sensitivity: 83.33\% \cite{13} \\
\hline
Attention Gate-based Dual-pathway Network (AGNet) & Coarse-to-fine framework with context and edge pathways. & Spine X-ray images, vertebrae dataset & Spinal diagnosis systems & Superior performance compared to state-of-the-art methods \cite{14} \\
\hline
Selective Binary Gaussian Filtering Regularized Level Set & Fully automatic segmentation with morphological operations. & 10 trauma patient datasets & Diagnosis, therapy, surgical intervention & DSC: 90.86\% (whole spine), 86.08\% (thoracic), 95.61\% (lumbar) \cite{15} \\
\hline
2D Centroid-detection Guidance Segmentation Network (CD-VerTransUNet) & Utilizes a global information and multitasking approach. & VerSe’20, scoliotic dataset & Diagnosis, treatment planning & DSC: 75.15\% (sagittal), 71.16\% (coronal) \cite{pereanez2015accurate} \\
\hline
Atlas-based Segmentation & Registers multiple atlases to target data and use label fusion. & Training dataset from MICCAI workshop & Not specified & Average DICE score: 0.94 \cite{17} \\
\hline
\end{tabular}}}
\label{tab:vertebra_segmentation_methods}
\end{table*}

In \citep{lessmann2018iterative}, a two-stage iterative technique was employed: lower-resolution vertebrae were first sequentially identified and segmented, then low-resolution masks were refined through a CNN. This technique led to a single phase fully convolutional network in \citep{lessmann2019iterative}. In \citep{cheng2017automatic} the authors propose a pixel based segmentation for 2D patch based pixel classifications. In their work, a deep CNN model based on the Active Appearance Model (AAM) is used to aid the initial level of the pipeline by providing a rough bounding area, which is then passed through a modification process from the Atlas-based AAM at the secondary level.  Finally, in \citep{payer2020coarse}, purposely built fully convolutional networks were employed in a coarse-to-fine segmentation technique that comprised vertebra labeling, spine localization, and vertebrae segmentation. 
Unet models typically adopt an asymmetrical U-shaped structure \citep{lu2023lumbar} that achieves good results in medical segmentation tasks, however, they are limited by their skip layer connections which constrain the encoder and decoder to perform feature fusion only within layers of the same depth. As a result, specific details are lost due to the inability to fuse semantic information with different scales. The structure of the basic Unet is optimized in \citep{zhou2018unet++} (Unet ++), which utilizes an efficient collection of Unets of different depths to alleviate the issue of the unknown network depth. Furthermore, Unet++ recreates skip connections to aggregate features of different semantic scales at the decoder subnetworks and offers a pruning method to accelerate inference speed \citep{lu2023lumbar}. This is further improved in Unet 3+ \citep{huang2020unet}, which employs deep supervision and full-scale skip connections. To overcome the confusion sparked by similarities between MRI scans of different disks, the authors in \citep{hille2018vertebral} used bounding boxes to crop the estimated center, employed zero padding and fed them into the convolutional network. ReLU activation function is used in the 3D convolutional network with the addition of a dropout operation. In \citep{payer2020coarse}, a 3D Unet is first used to roughly localize the spine, followed by a heat map regression for spine localization and identification using the Spatial Configuration Net \citep{payer2019integrating}, and then finally, a 3D Unet is used for binary segmentation on the identified vertebrae. In \citep{sekuboyina2017attention}, patch-based binary segmentation was executed, then heatmaps of vertebrae masks were denoised. In \citep{sekuboyina2017localisation} a simple multi-layer perceptron was first trained to regress the localization of the region of the lumbar spine, then a Unet was trained towards multiclass segmentation. The multi-layer perceptron in \citep{sekuboyina2017attention} was then substituted in \citep{janssens2018fully} with two sequential CNNs for multiclass segmentation. A recent study demonstrated that nnUnet can be embedded within a universal model to enable the segmentation of 33 different anatomical structures including vertebrae, pelvic bones, and abdominal organs from CT images~\citep{liu2022universal}. Experimental results using VerSe'19 and VerSe'20 datasets demonstrate high performance segmentation in terms of Dice coefficient and Hausdorff distance as shown in~\citep{liu2022universal} (Table 5).

SSAE is a dual-stage encoder-decoder architecture in which pixel intensities are first encoded in the encoder through low-dimensional features, which are later used to reconstruct the original intensities in the decoder. SSAE is a fully connected network that represents weight features in a single global weight matrix, unlike CNN, which is more oriented towards partial connections and better emphasizes the significance of locality.

\begin{figure*}
\centering
\includegraphics[width=0.9\textwidth]{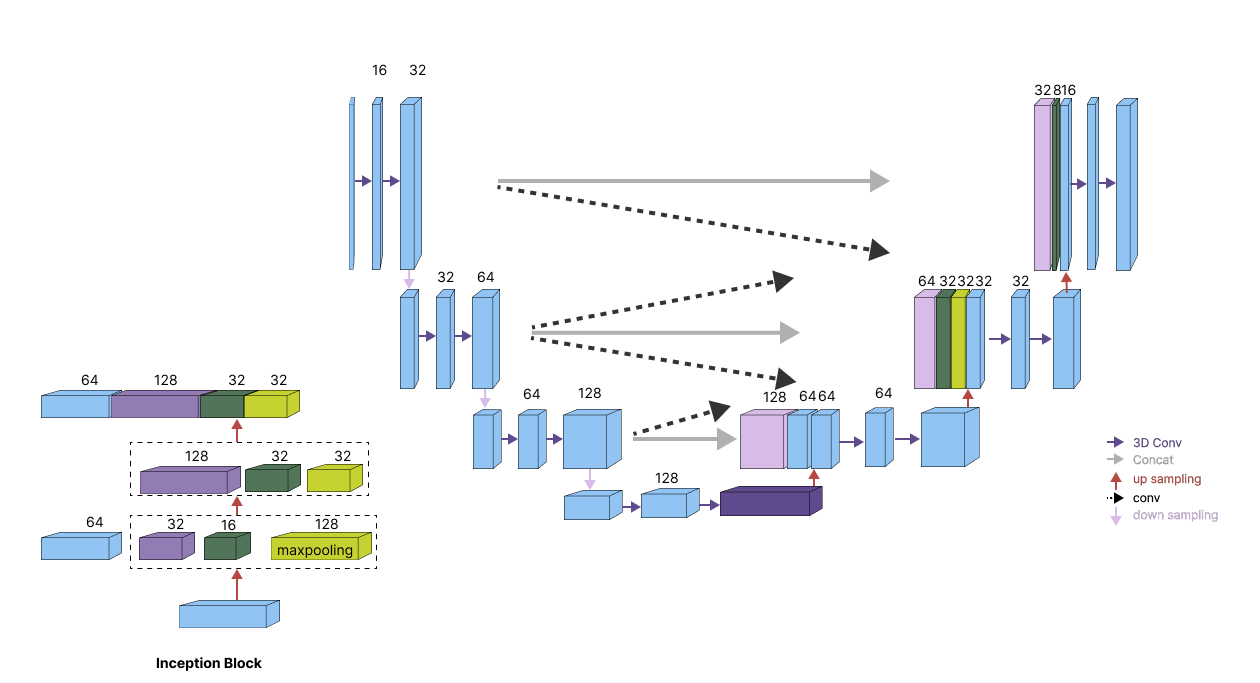}
\caption{3D X Unet architecture in \citep{lu2023lumbar}.}
\label{R3_fig30}
\end{figure*}

\begin{figure*}
\centering
\includegraphics[width=0.9\textwidth]{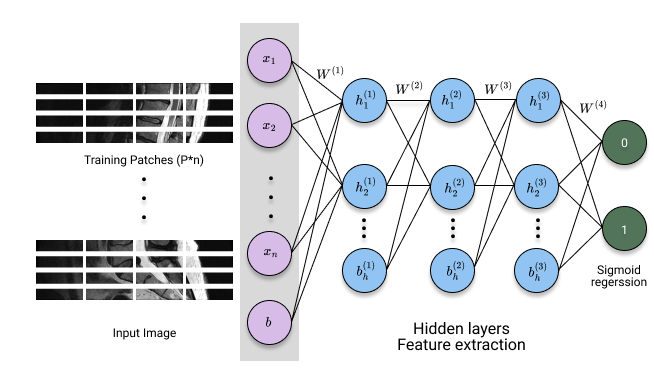}
\caption{SSAE architecture in \citep{qadri2023ct}.}
\label{R3_fig31}
\end{figure*}

The first instance of SSAE employment in MIS can be traced back to \citep{hinton2006reducing}, in which the first deep autoencoder network was developed. In \citep{shin2012stacked}, stacked autoencoders were employed in MRIs to identify organs and in \citep{aslam2021breath} a CAD system employed SSAE to identify gastric cancer from breath samples. In \citep{praveen2018ischemic} an SVM classifier was employed to receive the output from SAE layers that segment stroke lesions, and in \citep{qadri2019vertebrae} SSAE was used to create a model for vertebral segmentation. In \citep{wang2019automatic} the authors achieved automatic vertebrae localization and identification using SSAE and a structured regression forest, and in \citep{li2019longitudinal} a SSAE framework was proposed to diagnose Parkinson's disease. SSAE was favored over CNN in \citep{qadri2023ct} owing to its unsupervised extraction of high-level features from the bottom up, leading to more efficient representations, precise patch classifications, and a more robust CT vertebral segmentation. They used a PE module to extract overlapping image patches and their labeling with predefined pixel ratios, whereas a RUS module was employed to address the class imbalance problem. Figures~\ref{R3_fig30} and \ref{R3_fig31} represent sample Unet-based \citep{lu2023lumbar} and SSAE-based \citep{qadri2023ct} architectures, respectively. A summary of these methods is listed in Table~\ref{tab:vertebra_segmentation_methods}.

\section{Conclusion}
Medical image segmentation remains a profoundly active and vital research area, driven by its direct impact on improving patient diagnostics and therapeutic outcomes. As we have explored throughout this survey, the journey of the transformation of advanced computational tools. This paper has served as a comprehensive guide, thoroughly outlining the evolution of methodologies, from traditional approaches like thresholding and region-based methods to the latest deep learning paradigms such as U-Nets and Transformers. We aimed to equip future researchers with a robust toolkit for developing more sophisticated and clinically relevant models that adapt to the unique characteristics of diverse imaging modalities like X-ray, CT, and MRI.
The ongoing challenges in medical image segmentation, including data limitations, inherent noise and artifacts, and the exploration for universally applicable methods, underscore the extended potential for further innovation. Our survey particularly highlights the importance of distributed learning, especially federated learning, as a critical avenue for future exploration, given the increasing emphasis on patient data privacy and secure multi-institutional collaboration. Furthermore, the rigorous assessment of uncertainty quantification methods alongside lightweight models ensures that deep learning solutions are not only accurate but also trustworthy, interpretable, and deployable in the varied, resource-constrained clinical settings where they are most needed.

\section*{Acknowledgment}
This work was supported by JST, PRESTO Grant Number JPMJPR23P7, Japan.


\bibliographystyle{cas-model2-names}

\bibliography{sn-bibliography}



\end{document}